\documentclass[10pt,twocolumn,letterpaper]{article}

\usepackage[pagenumbers]{cvpr} % To force page numbers, e.g. for an arXiv version

\usepackage[dvipsnames]{xcolor}

\definecolor{cvprblue}{rgb}{0.21,0.49,0.74}
\usepackage[pagebackref,breaklinks,colorlinks,citecolor=cvprblue]{hyperref}
\usepackage{multirow}
\usepackage{multicol}
\usepackage{adjustbox}
\usepackage{algorithm}
\usepackage{algorithmic}
\usepackage{colortbl}

\def\paperID{16399} % *** Enter the Paper ID here
\def\confName{CVPR}
\def\confYear{2024}

\title{\ours: Improving Optical Flow Estimation by \\ Occlusion and Consistency Aware Interpolation }
\author{
Jisoo Jeong~~~
Hong Cai~~~
Risheek Garrepalli~~~
Jamie Menjay Lin~~~
Munawar Hayat~~~
Fatih Porikli~~~
\smallskip
\\
Qualcomm AI Research$^{\dagger}$~~~
\\
\smallskip
{\tt\small\{jisojeon, hongcai, rgarrepa, jmlin, hayat, fporikli\}@qti.qualcomm.com \vspace{-12pt}}
}

\newcommand{\ignore}[2]{\hspace{0in}#2}

\usepackage{xspace}
\newcommand{\ours}{{OCAI}\xspace}

\begin{document}
\maketitle

\begin{abstract}
\vspace{-8pt}

The scarcity of ground-truth labels poses one major challenge in developing optical flow estimation models that are both generalizable and robust. While current methods rely on data augmentation, they have yet to fully exploit the rich information available in labeled video sequences. We propose \ours, a method that supports robust frame interpolation by generating intermediate video frames alongside optical flows in between. Utilizing a forward warping approach, \ours employs occlusion awareness to resolve ambiguities in pixel values and fills in missing values by leveraging the forward-backward consistency of optical flows. Additionally, we introduce a teacher-student style semi-supervised learning method on top of the interpolated frames. Using a pair of unlabeled frames and the teacher model's predicted optical flow, we generate interpolated frames and flows to train a student model. The teacher's weights are maintained using Exponential Moving Averaging of the student. Our evaluations demonstrate perceptually superior interpolation quality and enhanced optical flow accuracy on established benchmarks such as Sintel and KITTI.
\vspace{-10pt}

\end{abstract}    
\begin{figure}[t!]
    \centering
    \includegraphics[width=1.0\linewidth]{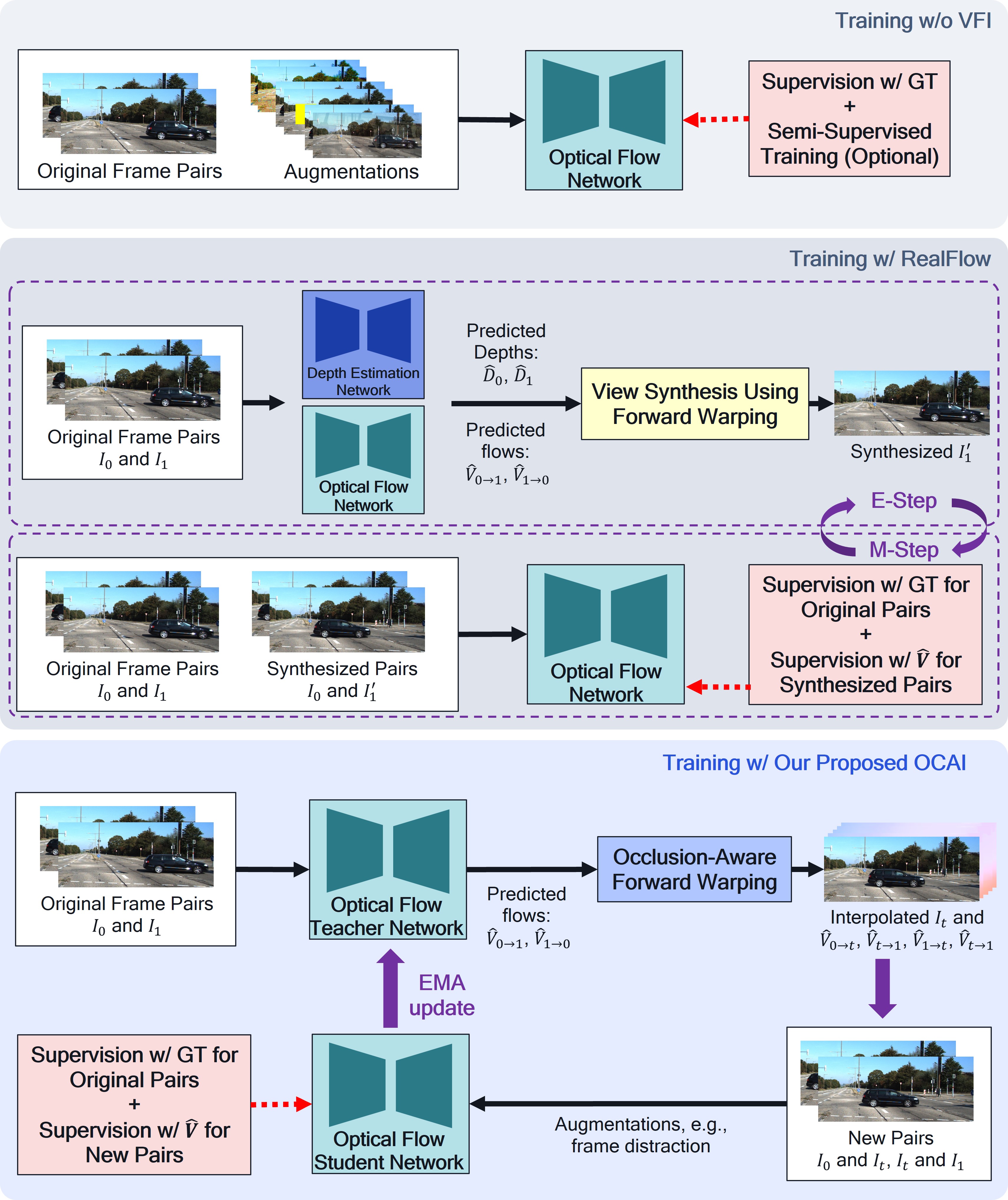} 
    \vskip -8pt 
    \caption{\small \textbf{Top:} Many existing data augmentation approaches focus on modifying the existing frames ~\cite{liu2019selflow, teed2020raft, jeong2023distractflow}. 
    \textbf{Middle:} While RealFlow \cite{han2022realflow} employs optical flow model to generate new frames and iteratively update the model with generated frames, it requires compute expensive steps of depth estimation and EM optimization. 
    % A recent work, RealFlow~\cite{han2022realflow} synthesizes new frames based on the network's predicted optical flow, which are used to iteratively update the network during training (middle). 
    % \mh{lets mention limitation of RealFlow here. can we replace last sentence with something like --> While RealFlow \cite{han2022realflow} employs optical flow model to generate new frames and iteratively update the model with generated frames (middle), it requires compute expensive steps of depth estimation and EM optimization.} 
    \textbf{Bottom:} Our proposed \ours allows flexible, robust video interpolation at any intermediate time step, and leverages interpolated frames and flows to efficiently train the model in a semi-supervised setting, significantly improving optical flow estimation.}
    \label{fig:teaser}
    \vspace{-10pt}
\end{figure} 

\section{Introduction}
\label{sec:intro}
\vspace{-3pt}

{\let\thefootnote\relax\footnotetext{{
\hspace{-6.5mm} $\dagger$ Qualcomm AI Research is an initiative of Qualcomm Technologies, Inc.}}}

Optical flow estimation and Video Frame Interpolation (VFI) share a complementary relationship. Accurate optical flow contributes significantly to various downstream tasks such as video compression~\cite{wu2018video, lu2019dvc}, video denoising and blur removal~\cite{buades2016patch, yuan2020efficient, zhang2021deep}, action recognition~\cite{lee2018motion, cai2019temporal}, and VFI stands as one of these applications. Pixel-level correspondence by optical flow enables estimating pixel-level movement and generating intermediate frames (or inter-frames). While utilizing flow-based methods is a common practice in VFI~\cite{kong2022ifrnet, lu2022video, huang2022real, li2023amt}, leveraging inter-frames to train optical flow models is relatively less explored.

%Optical flow estimation and Video Frame Interpolation (VFI) have a complementary relationship. Accurate estimation of optical flow can help multiple downstream tasks, e.g., video compression~\cite{wu2018video, lu2019dvc}, action recognition~\cite{lee2018motion, cai2019temporal}, and VFI is one of them. Pixel level correspondence by optical flow enables estimating pixel level movement and generating intermediate frames (or inter-frames).
% For this reason, flow-based methods are in the mainstream of VFI~\cite{kong2022ifrnet, lu2022video, huang2022real, li2023amt}. On the other hand, there is a possibility of using the interpolated frames to improve the optical flow model training in a semi-supervised manner, which is a less explored area. 
%While flow-based methods are a common practice in VFI~\cite{kong2022ifrnet, lu2022video, huang2022real, li2023amt}, leveraging inter-frames for training optical flow models is yet less explored.

While the scarcity of ground-truth data has long been a critical challenge in learning-based optical flow estimation~\cite{dosovitskiy2015flownet, ilg2017flownet, ranjan2017optical, sun2018pwc, teed2020raft, huang2022flowformer}, there has been little attention into leveraging Video Frame Interpolation (VFI) to augment the training of optical flow networks. State-of-the-art VFI models predominantly employ deep networks trained to interpolate the exact middle frame between two consecutive time steps within a video sequence, which restricts their capability to generate frames at other intermediate time instances thus hampering their ability to produce optical flows between existing and intermediate frames reliably. Besides, these models lack generalizability across new domains without necessitating finetuning or retraining. These limitations hinder the potential use of existing VFI models for augmenting the training data for optical flow models.
%While the lack of ground-truth data has been a known, critical challenge for learning-based optical flow estimation~\cite{dosovitskiy2015flownet, ilg2017flownet,ranjan2017optical, sun2018pwc, teed2020raft, huang2022flowformer}, there is little attention to using VFI to enhance the training of optical flow networks. State-of-the-art (SOTA) VFI models predominately utilize deep neural networks, which are typically trained to interpolate the exact middle frame between two consecutive time steps in an existing video sequence. This makes it inflexible for them to generate frames at other intermediate time instances. In addition, these models cannot reliably produce the optical flows between the intermediate frame and the existing frames. Furthermore, they do not generalize well to new domains without finetuning or retraining. These factors limit the usage of existing VFI models for augmenting the training data for optical flow models.

To address data scarcity in optical flow training: data augmentation \cite{sun2021autoflow, jeong2023distractflow}, data generation \cite{dosovitskiy2015flownet, sun2021autoflow, han2022realflow}, and semi-supervised learning \cite{jeong2022imposing, im2022semi, han2022realflow,jeong2023distractflow} have been explored. Most of the current data augmentation methods prioritize modifying the existing frames, e.g.,~\cite{jeong2023distractflow}, as illustrated in Fig.~\ref{fig:teaser} (top). Notably, RealFlow\cite{han2022realflow} stands out by synthesizing a new second frame via forward warping. Given a pair of frames and the model's prediction, the original first and second frames form a new training pair, used to update model weights. Model prediction, frame synthesis, and model update are iterated in an Expectation-Maximization (EM) framework, as depicted in Fig.~\ref{fig:teaser} (middle). While it is possible to use forward warping to interpolate frames, RealFlow focuses only on making the model's prediction consistent with frame synthesis. Moreover, the EM steps significantly increase the training workload.

In this paper, we introduce \ours, a novel approach for training optical flow networks within a semi-supervised framework using readily available unlabeled pairs. Our method leverages video frames and flow interpolation to achieve this goal. \ours implements an occlusion-aware forward warping technique, facilitating the interpolation of both inter-frames and intermediate flows, and can effectively tackle pixel value ambiguities during the warping process without needing any depth information and can do confidence-aware estimation of \emph{missing values} utilizing forward-backward consistency. Our algorithm can perform interpolation at any intermediate time step, thus offering a broad diversity of data and motion ranges essential for training optical flow models. Our innovative teacher-student style semi-supervised learning scheme utilizes these resulting inter-frames to train optical flow networks effectively.

In summary, our main contributions are as follows:
\begin{itemize}

\item We propose a novel approach, \ours, that tackles the data scarcity challenge in training optical flow models, by exploiting useful, hidden information in existing videos. Specifically, we interpolating frames and flows to generate supplementary data, and use them in a semi-supervised learning framework.

\item To do this, we propose a new, effective video interpolation method that derives occlusion information to address ambiguous pixels and fill in holes by exploiting optical flow consistency. Our algorithm flexibly generates high-quality intermediate frames and reliable intermediate flows along with corresponding confidence maps at any intermediate time step.

% \vspace{-0.5mm}
\item We devise a new teacher-student semi-supervised learning strategy leveraging VFI to train an optical flow network, incorporating exponential moving averaging (EMA) to enhance training stability.% and performance.  

% \vspace{-0.5mm}
\item 
% We demonstrate the efficacy of our proposed MoreFlow on standard benchmarks like Sintel and KITTI, and achieve high-quality video interpolation compared with the existing SOTA. 
We demonstrate that \ours achieves higher quality in video interpolation than existing SOTA methods on standard datasets including Sintel and KITTI.
By incorporating interpolated video information, our semi-supervised learning scheme significantly improves the optical flow estimation performance, e.g., 0.5+ Fl-all reduction on KITTI test set when comparing to latest SOTA.

\end{itemize}

\begin{figure*}[t!]
\centering
\includegraphics[width=0.95\linewidth]{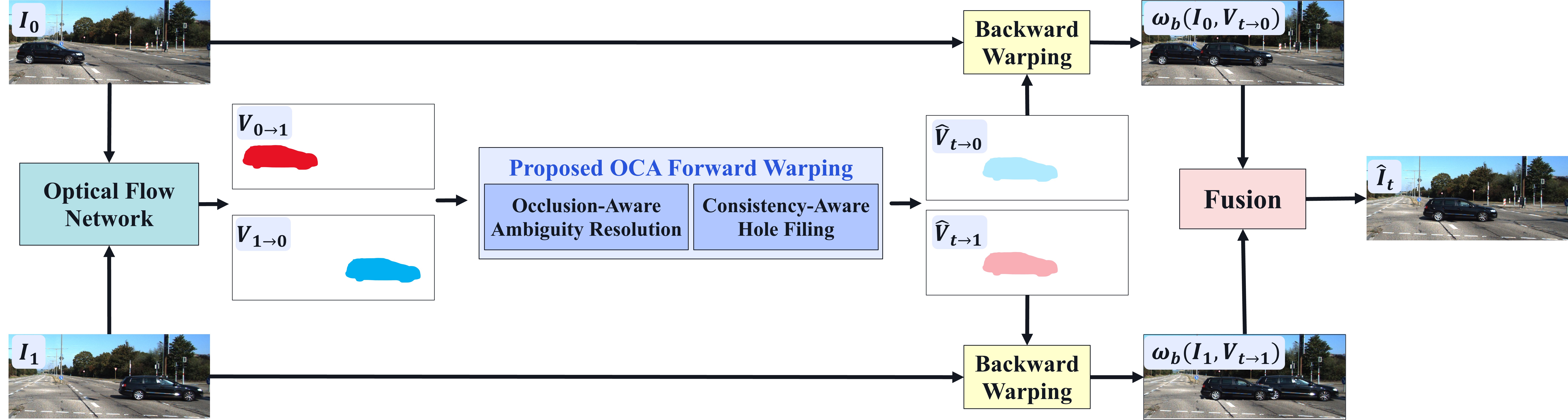}
\vspace{-6pt}
\caption{\small Our proposed video interpolation algorithm using occlusion and consistency-aware forward warping.}
\label{fig:intro_vfi}
\vspace{-8pt}
\end{figure*}

\section{Related Work and Preliminaries}
\label{sec:related}
\vspace{-3pt}

% Occlusion mask ($O$) is obtained by forward-backward consistency \cite{meister2018unflow}
% \begin{equation} \label{For-Back_Con}
% \begin{split}
% |\widehat{V}^{f}(x) + \widehat{V}^{b}(x+\widehat{V}^f(x))|^2 \quad \quad \quad \quad \quad \quad \quad \\
% < \gamma_{1} \Big{(}|\widehat{V}^{f}|^2 + |\widehat{V}^{b}(x+\widehat{V}^f(x))|^2 \Big{)} + \gamma_{2},
% \end{split}
% \end{equation}
% where $\gamma_1$ = 0.01 and $\gamma_2$ = 0.5 from~\cite{meister2018unflow}.

% Forward Warping for Video 
Consider two consecutive video frames, $I_0$ and $I_1$. We denote the optical flow from $I_{0}$ to $I_{1}$ as $V_{0 \rightarrow 1}$ and the inter-frame as $I_{t}$, where $t \in (0,\,1)$. 
% The occlusion map for and occlusion-aware mask for $I_{0}$ are denoted as $O_{0}$ and $M_{0}$, respectively.
We use $\omega_{b}$ and $\omega_{f}$ to denote backward and forward warping operations, respectively. For instance, backward warped image $\widehat{I}^{b}_0$ can be obtained by $\omega_{b}(I_{1},V_{0 \rightarrow 1})$ and forward warped image $\widehat{I}^{f}_1$ can be computed by $\omega_{f}(I_{0},V_{0 \rightarrow 1})$.
% $D_{\lambda}(I_{t}, \tilde{I}_d)$ is the perturbed image of $I_t$ obtained by mixing with another image $\tilde{I}_d$ ($\lambda \in (0,\,1)$). $D_{\lambda}(I_{t}, \tilde{I}_d)$ is computed the same as defined in \cite{jeong2023distractflow}.

\subsection{Video Frame Interpolation}
\vspace{-3pt}
% Several deep learning based approaches have been proposed for Video Frame Interpolation \cite{peleg2019net, cheng2020video, choi2020channel, shi2022video, bao2019depth, bao2019memc, kong2022ifrnet, li2023amt}. 

Video Frame Interpolation (VFI) algorithms can be divided into three categories: flow-based~\cite{bao2019depth, bao2019memc, kong2022ifrnet, li2023amt}, kernel-based~\cite{peleg2019net, cheng2020video} and hallucination-based~\cite{choi2020channel, shi2022video} approaches. 
Among these, flow-based methods have shown SOTA performance~\cite{kong2022ifrnet, li2023amt}. These methods predict optical flows, and apply forward or backward warping using existing frames and the flows. The warped images are fused with residual information computed by the deep learning network. 

% Video Frame Interpolation (VFI) algorithms can be divided into three categories: flow-based~\cite{bao2019depth, bao2019memc, kong2022ifrnet, li2023amt}, kernel-based~\cite{peleg2019net, cheng2020video} and hallucination-based~\cite{choi2020channel, shi2022video} approaches. 
% Among these, flow-based VFI approaches have shown SOTA performance thanks to optical flow robustness \cite{kong2022ifrnet, li2023amt}. In Flow-based approach, it predicts optical flows and applies forward or backward warping using source images and optical flows. And then, it combines warped images with residual information computed by network. 

% In recent years, optical flow based Video Frame Interpolation (VFI) methods generate accurate inter-frame images. 

\vspace{3pt}
\noindent \textbf{Backward-warping-based VFI:} 
% \subsubsection{Backward Warping based VFI methods}
The VFI network computes two optical flows, $V_{t \rightarrow 0}$ and $V_{t \rightarrow 1}$, a weighting mask (M), and the residual (R) using the two frames, $I_{0}$ and $I_{1}$, where $t$ is an intermediate time step. It applies backward warping to $I_{0}$ and $I_{1}$, combines them using the mask, and finally, adds the residual information, as follows: \vspace{-5pt}
\begin{equation} \label{real_interframe}
\widehat{I}_{t}\! =\! M \cdot \omega_{b}(I_{0}, V_{t \rightarrow 0}) + (1\!-\!M) \cdot  \omega_{b}(I_{1}, V_{t \rightarrow 1}) + R,
\vspace{-5pt}
\end{equation}
where $\widehat{I}_t$ is the interpolated frame. This approach requires predicting $V_{t \rightarrow 0}$ and $V_{t \rightarrow 1}$ without $I_{t}$, which is challenging when there are large displacements.

% softmax splatting. 
% RealFlow
% dd
% dd
% \subsubsection{Forward Warping based VFI methods}
% \subsection{Forward Warping Using RealFlow~\cite{han2022realflow}}
% RealFlow generates a new dataset using predicted optical flow and new generated frame between $I_0$ and $I_1$. 
\vspace{3pt}
\noindent \textbf{Forward-warping-based VFI:} 
Softmax Splatting~\cite{niklaus2020softmax} is introduced for forward warping. It predicts two optical flows, $V_{0 \rightarrow 1}$ and $V_{1 \rightarrow 0}$, between $I_0$ and $I_1$, and assumes that $V_{0 \rightarrow t} = t \cdot V_{0 \rightarrow 1}$ and $V_{1 \rightarrow t} = (1-t) \cdot V_{1 \rightarrow 0}$. 
% generates an inter-frame image $I_t$ using depth-based Softmax Splatting \cite{niklaus2020softmax} method.
It generates $I_{t}$ with the following steps:\vspace{-5pt}
\begin{equation} \label{real_interframe}
\begin{split}
\text{Let }  & u = p - (q +  V_{0 \rightarrow t}), \\
 & b(u) = \text{max}(0,\, 1-|u_x|) \cdot \text{max}(0,\, 1-|u_y|).
\end{split}
\end{equation}
\begin{equation} \label{real_interframe3}
\text{Then, } \widehat{I}_t(p) = 
\frac{\Sigma_{q} \text{exp}(Z_{0}(q)) \cdot I_{0}(q) \cdot b(u)}{\Sigma_{q} \text{exp}(Z_{0}(q)) \cdot b(u)}, \quad \quad \
\end{equation}
where $p$ is target grid index, $q$ is source grid index, $V_{0 \rightarrow t}$ is the optical flow between $I_0$ and $I_t$, $u_x$ and $u_y$ are x and y components of $u$, $b(u)$ indicates the mapping between pixels in $I_0$ and $I_t$, i.e., $b(u)\!=\!1$ if $p$ and $q$ map to each other. 

Softmax Splatting uses a learnable weighting $Z$ and Realistic Image Pair Rendering (RIPR) in RealFlow~\cite{han2022realflow} uses the inverse depth map of $I_{0}$ from a monocular depth estimation model. 
In forward warping, there can be pixel ambiguities, i.e., two source pixels can be mapped to the same target pixel location. 
In RIPR of RealFlow, as the weighting is based on depth, source pixels closer to the camera are chosen to resolve ambiguities. However, this requires additional computation, e.g., RIPR uses a depth network with $>$300M parameters, and is susceptible to depth estimation errors. 
In addition, RIPR introduced the Bi-directional Hole Filling which fills in the holes in $w_{f}(I_{0}, V_{0 \rightarrow t})$ image using $w_{f}(I_{1}, V_{1 \rightarrow t})$ values.
In OCAI, we introduce an occlusion-aware forward-warping algorithm that can generate accurate inter-frame without depth estimation.
% generates two forward warped images ($w_{f}(I_{0}, V_{0 \rightarrow t})$, $w_{f}(I_{1}, V_{1 \rightarrow t})$) and then fill in the $w_{f}(I_{1}, V_{1 \rightarrow t})$ values in $w_{f}(I_{0}, V_{0 \rightarrow t})$ hole. }

\subsection{Semi-Supervised Optical Flow Model Training}
\vspace{-3pt}
% Most of semi-supervised optical flow methods have adopted classification semi-supervised training approaches. RAFT-OCTC \cite{jeong2022imposing} introduced transformation consistency \cite{laine2016temporal, tarvainen2017mean} to regression task. DistractFlow~\cite{jeong2023distractflow} introduces a new data augmentation method for optical flow estimation inspired by Interpolation regularization\cite{zhang2017mixup, verma2019interpolation, jeong2021interpolation} and Fixmatch \cite{sohn2020fixmatch}. DistractFlow also proposes forward-backward consistency \cite{meister2018unflow} based confidence map, which enables a semi-supervised training scheme. The confidence map computation is as follows:
% Most semi-supervised optical flow methods are inspired by semi-supervised classification. \hc{What does this sentence mean? []} 
RAFT-OCTC~\cite{jeong2022imposing} introduces transformation consistency~\cite{laine2016temporal, tarvainen2017mean} to regressing optical flows. 
FlowSupervisor~\cite{im2022semi} proposes a new teacher network that can be trained stably. It uses the same encoder weights for both teacher and student, and computes the loss over all the pixels without using confidence masking.
DistractFlow~\cite{jeong2023distractflow} introduces semantic augmentation which is inspired by Interpolation Regularization~\cite{zhang2017mixup, verma2019interpolation, jeong2021interpolation} and FixMatch~\cite{sohn2020fixmatch}.  In particular, DistractFlow proposes to use a confidence map derived from forward-backward consistency~\cite{meister2018unflow}, which improves the stability of semi-supervised training. The confidence map is computed as follows: \vspace{-5pt}
\begin{equation} \label{Con_map}
% \begin{split}
C_{0,1}\! =\! \text{exp}  {\Bigg{(}\!\!\!-\!\frac{|\widehat{V}_{0 \rightarrow 1}(x)\! +\! \widehat{V}_{1 \rightarrow 0}(x\!+\!\widehat{V}_{0 \rightarrow 1}(x))|^2}{\gamma_{1} (|\widehat{V}_{0 \rightarrow 1}|^2\! +\! |\widehat{V}_{1 \rightarrow 0}(x\!+\!\widehat{V}_{0 \rightarrow 1})|^2 )\! + \!\gamma_{2} } \Bigg{)}},
% \end{split}
\vspace{-2pt}
\end{equation}
where $\gamma_1$ = 0.01 and $\gamma_2$ = 0.5 from~\cite{meister2018unflow}.

RealFlow~\cite{han2022realflow} uses Expectation-Maximization (EM) to train an optical flow model in a semi-supervised setting. It first trains the model using supervision from existing ground-truth data. Then, it synthesizes new data based on the predicted optical flows and forward warping. After that, it trains the network with the new data. RealFlow repeats the these steps several times in the training process, which is computationally expensive.

Many semi-supervised learning algorithms~\cite{tarvainen2017mean, liu2021unbiased} for other tasks, such as classification and object detection, employ Exponential Moving Average (EMA) to robustly and stably update the teacher network, using a temporal ensemble of the student network. In this paper, we leverage video interpolation and several semi-supervised training techniques to enhance model performance.
% Therefore, we also employ EMA for robust semi-supervised training.

\section{Proposed Approach}
\label{sec:method}
\vspace{-3pt}
\ours improves the accuracy of optical flow models by generating diverse, high-quality intermediate frames and flows, and trains the network with new pairs in a semi-supervised learning framework. 
In Section~\ref{sec:method_vfi}, we present our occlusion and consistency aware forward warping algorithm for video interpolation. 
% In Section~\ref{sec:method_vfi}, we present our video interpolation algorithm using Occlusion- and Consistency-Aware (OCA) forward warping.
Next, in Section~\ref{sec:method_self}, we propose a semi-supervised learning strategy to leverage video interpolation to better train optical flow networks. 

% Our strategy of improving the accuracy of optical flow model is that generates more accurate inter-frame images and trains the optical flow model with new generated data pairs which are applied Distract noise. In Section~\ref{sec:method_vfi}, we describe how we generate the accurate inter-frame image using occlusion information. Next, in Section~\ref{sec:method_self}, we leverage realistic data augmentation method to our new generated pairs in a supervised and semi-supervised manner. 

% \begin{figure*}[ht]
% \centering
% \includegraphics[width=0.95\linewidth]{img/Fig_occ_vfi_module.png}
% % \vspace{-10pt}
% \caption{Occlusion aware frame interpolation module }
% % \vspace{-10pt}
% \label{fig:occ_module}
% % \vspace{-5mm}
% \end{figure*}

% We can focus on just describing the procedure here
\subsection{Occlusion and Consistency Aware Interpolation Using Forward Warping (OCAI)}
\label{sec:method_vfi} \vspace{-3pt}

% % Forward Warping for Video 
% Consider two consecutive video frames, $I_0$ and $I_1$. We denote the optical flow from $I_{0}$ to $I_{1}$ as $V_{0 \rightarrow 1}$ and the inter-frame as $I_{t}$, where $t \in (0,\,1)$. 
% % The occlusion map for and occlusion-aware mask for $I_{0}$ are denoted as $O_{0}$ and $M_{0}$, respectively.
% We use $\omega_{b}$ and $\omega_{f}$ to denote backward and forward warping operations, respectively. For instance, backward warped image $\widehat{I}^{b}_0$ can be obtained by $\omega_{b}(I_{1},V_{0 \rightarrow 1})$ and forward warped image $\widehat{I}^{f}_1$ can be computed by $\omega_{f}(I_{0},V_{0 \rightarrow 1})$. 
% % $D_{\lambda}(I_{t}, \tilde{I}_d)$ is the perturbed image of $I_t$ obtained by mixing with another image $\tilde{I}_d$ ($\lambda \in (0,\,1)$). $D_{\lambda}(I_{t}, \tilde{I}_d)$ is computed the same as defined in \cite{jeong2023distractflow}.

Our goal is to generate an accurate estimation of the intermediate frame $I_t$ by using $I_0$, $I_1$, $V_{0\rightarrow1}$, and $V_{1\rightarrow0}$. In order to do this, we first need to estimate $V_{t\rightarrow0}$ and $V_{t\rightarrow1}$, based on which we can perform backward warping from $I_0$ and $I_1$ and fuse the warped images, $\omega_b(I_0, \widehat{V}_{t\rightarrow0})$ and $\omega_b(I_1, \widehat{V}_{t\rightarrow1})$, to generate the estimated inter-frame, $\widehat{I}_t$. 
In addition to generating $I_{t}$, we compute two intermediate optical flows, $V_{t \rightarrow 0}$ and $V_{t \rightarrow 1}$. This has several advantages. 
First, $V_{t \rightarrow 0}$ and $V_{t \rightarrow 1}$ can be used in optical flow training. Specifically, these optical flows enable computing a confidence map via forward-backward consistency, which is important to self-/semi-supervised training. Second, using the confidence map allows us to more accurately fuse warped versions of $I_0$ and $I_1$ to produce $\widehat{I}_t$. Previous approach~\cite{han2022realflow} generates $\widehat{I}_{t}$ from $I_{0}$ and fills hole using content generated from $I_{1}$; however, there are discrepancies since pixels have moved between $I_0$ and $I_1$. Finally, our method can generate better background. For instance, RealFlow has missing values near image boundaries whereas we do not have this issue (see last two rows in Fig.~\ref{exp:qualitative_vfi}).  
% \hc{any insight why?}. \textcolor{blue}{Even RealFlow also apply hole filling, there are holes. (If Two wapred images have holes in both images, it cannot fill in the hole. In our case is the same. But, In RealFlow case, it is just 0, and makes it black. But, in our case, it is 0 optical flow. So, we can get average of $I_0$ and $I_1$ values.}

% In addition to generating only $I_{t}$, we also compute two intermediate optical flows, $V_{t \rightarrow 0}$ and $V_{t \rightarrow 1}$, from $V_{0 \rightarrow 1}$ and $V_{1 \rightarrow 0}$. Generating additional optical flows have three advantage for frame interpolation. First, it generates additional optical flows ($V_{t \rightarrow 0}$ and $V_{t \rightarrow 1}$) which can be used in optical flow training. These optical flow is enable to generate the forward-backward based confidence map, and helps to the self-supervised training. Second, using confidence map from two additional optical flow, it can fuse two image more accurately. In RealFlow, it generates $I_{t}$ from $I_{0}$ and fills hole region using generated image from $I_{1}$. However, there are discrepancies since things have moved in $I_1$. Third, our method can generate more reasonable background.

\begin{figure}[t]
\centering
\includegraphics[width=0.85\linewidth]{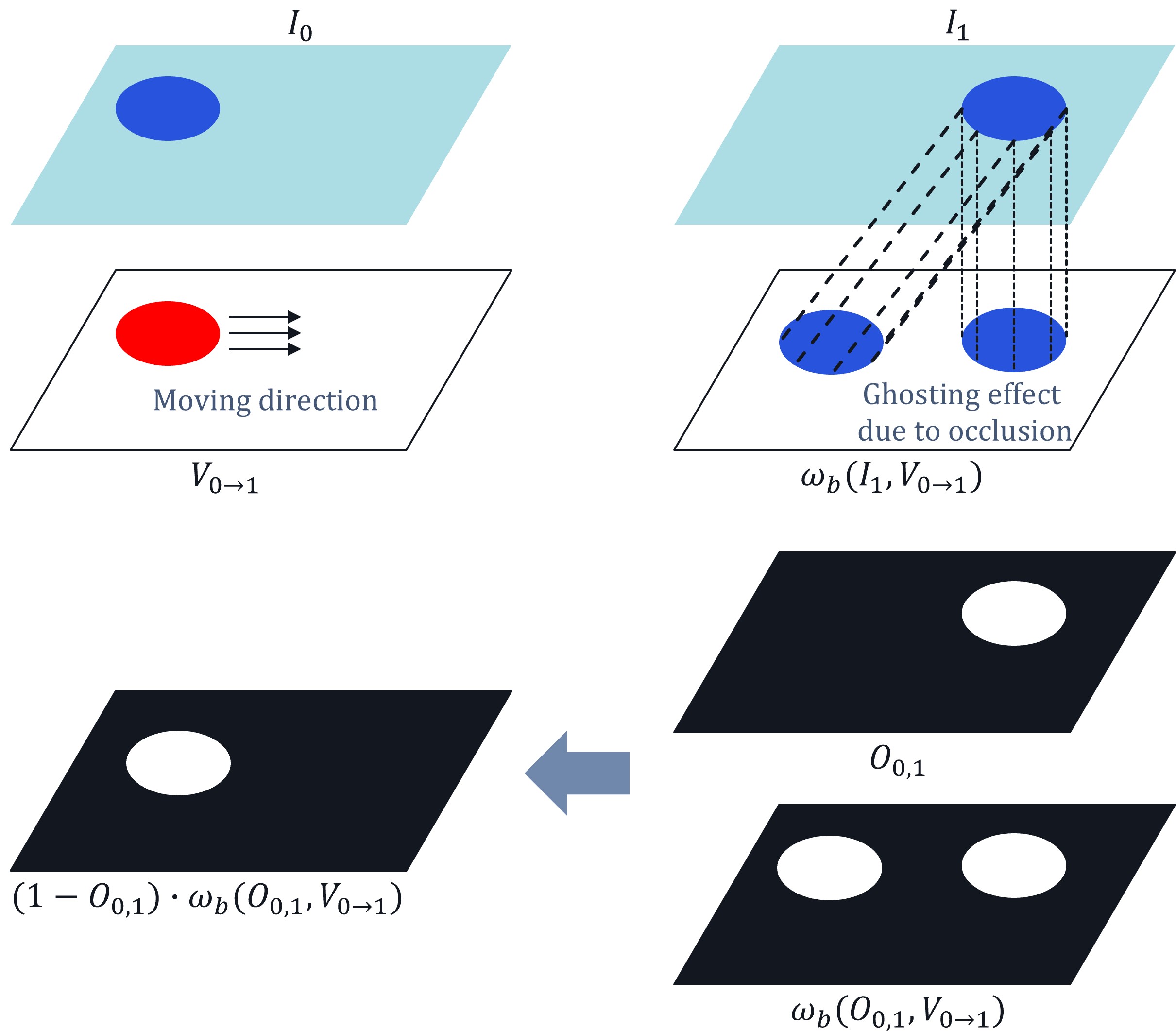}
\vspace{-6pt}
\caption{Visual illustration of deriving occlusion-aware weighting mask $M_0$. The top part shows backward warping with a moving object and static background. Ghosting effect happens since the background is occluded in $I_1$.
The bottom part illustrates how we derive the foreground mask on $I_0$, by performing backward warping and removing the ghosting effect. The foreground mask is used in the forward warping when synthesizing $\widehat{V}_{t\rightarrow1}$.}
\vspace{-8pt}
\label{fig:weighting_mask}
% \vspace{-5mm}
\end{figure}

Next, we present our OCAI in detail. 

\vspace{3pt}
\noindent \textbf{Generating Inter-Frame via Forward Warping}

First, we decompose $V_{0 \rightarrow 1}$ into $V_{0 \rightarrow t}$ and $V_{t \rightarrow 1}$:\vspace{-4pt}
\begin{equation} \label{decompose}
V_{0 \rightarrow 1}(x) = V_{0 \rightarrow t}(x) + V_{t \rightarrow 1}(x + V_{0 \rightarrow t}(x)). \\[-4pt]
\end{equation}

Assuming linear projected motion, $V_{0 \rightarrow t}$ can be properly approximated by $t \cdot V_{0 \rightarrow 1}$ and we have \vspace{-6pt}
\begin{equation} \label{assumption}
\begin{split}
V_{0 \rightarrow 1}(x) = t \cdot V_{0 \rightarrow 1}(x) + V_{t \rightarrow 1}(x + t \cdot V_{0 \rightarrow 1}(x)),\\[-8pt]
\end{split}
\end{equation}
\begin{equation} \label{assumption2}
\begin{split}
V_{t \rightarrow 1}(x + t \cdot V_{0 \rightarrow 1}(x)) = (1-t) \cdot V_{0 \rightarrow 1}(x).\\[-4pt]
\end{split}
\end{equation}

We see that by performing forward warping on $(1-t) \cdot V_{0 \rightarrow 1}(x)$ with optical flow $t \cdot V_{0 \rightarrow 1}(x)$, we can obtain $V_{t \rightarrow 1}(x)$, that is \vspace{-5pt}
\begin{equation} \label{forward_warp}
\begin{split}
V_{t \rightarrow 1}(x) = \omega_{f}((1-t) \cdot V_{0 \rightarrow 1}(x),\, t \cdot V_{0 \rightarrow 1}(x)).\\[-5pt]
\end{split}
\end{equation}

Similarly, we can compute $V_{t \rightarrow 0}(x)$ based on decomposing $V_{1\rightarrow0}$ and forward warping. 

By using $V_{t\rightarrow0}$ and $V_{t\rightarrow1}$, we can generate two versions of the intermediate frame $I_t$: $w_b(I_{0},V_{t \rightarrow 0})$ and $w_b(I_{0},V_{t \rightarrow 1})$, by backward warping. %$I_{t \leftarrow 0}$ and $I_{t \leftarrow 1}$
We can then use confidence maps, $C_{t,0}$ and $C_{t,1}$, to fuse them to estimate the inter-frame as follows: \vspace{-3pt}
% , where $\epsilon$ is a small number for numerical stability:
\begin{equation} \label{combine}
\small
\widehat{I}_{t} = \frac{C_{t,\,0}}{C_{t,\,0} + C_{t,\,1}} w_b(I_{0},V_{t \rightarrow 0}) + \frac{C_{t,\,1}}{C_{t,\,0} + C_{t,\,1} } w_b(I_{1},V_{t \rightarrow 1}),\\[-0pt]
% I_{t} = \frac{C_{t,\,0} + \epsilon}{C_{t,\,0} + C_{t,\,1} + 2\epsilon}\cdot w_b(I_{0},V_{t \rightarrow 0}) + \frac{C_{t,\,1} + \epsilon}{C_{t,\,0} + C_{t,\,1} + 2\epsilon}\cdot w_b(I_{1},V_{t \rightarrow 1}),
\end{equation}
where the confidence maps are calculated based on Eq.~\ref{Con_map}. For instance, $C_{t,\,0}$ is computed using $V_{t\rightarrow0}$ and $V_{0\rightarrow t}$.

In order to correctly perform forward warping, we need to resolve two issues: \textbf{pixel value ambiguity} and \textbf{missing pixel values}. 
Ambiguity is due to two pixels in the source frame moving to the same location in the target frame, in which case we need to understand which is closer to the camera and thus, should be chosen. Missing values is because a pixel location in the target frame can correspond to an object that is occluded in the source frame (and the occluding object moves away in target frame), where there are no pixels representing this occluded object. This can also be caused by an object moving closer to the camera and there are not enough pixels in the source frame to represent the object in the target frame.

We propose occlusion-aware weighting to resolve pixel value ambiguity and choose the pixel that corresponds to what is closer to the camera. We further introduce a hole-filling method based on the forward-backward consistency of optical flow. Fig.~\ref{fig:intro_vfi} provides an overview of our proposed forward warping approach.

% \newpage

\begin{figure*}[t]
\centering
\includegraphics[width=0.99\linewidth]{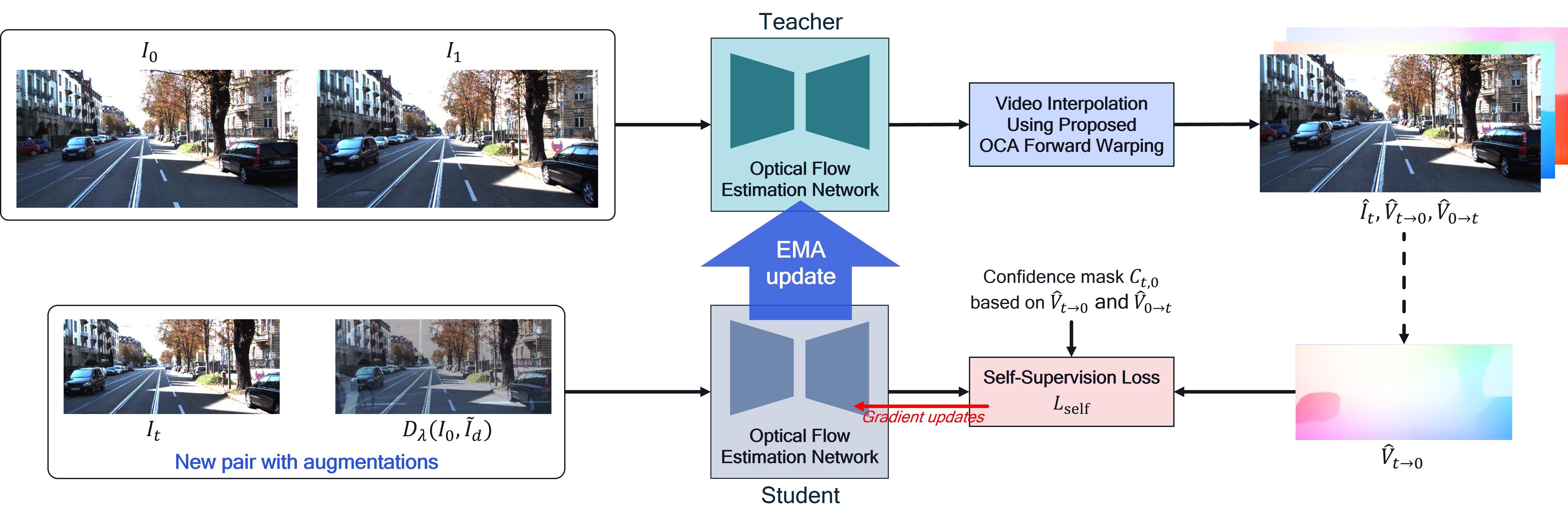}
\vspace{-11pt}
\caption{Self-supervision using interpolated video frames and flows in a teacher-student learning setting. Note that the student model is trained with both the self-supervision shown in the figure and the supervision from available ground-truth data.}
\vspace{-2pt}
\label{fig:overview}
% \vspace{-5mm}
\end{figure*}

\vspace{3pt}
\noindent \textbf{Occlusion-Aware Weighting to Resolve Ambiguity}

We resolve pixel value ambiguity via occlusion understanding. Specifically, we assume that \textit{when a pixel is not occluded but creates occlusion for other pixels, it corresponds to an object closer to the camera}. 
% Previous work~\cite{han2022realflow} uses depth to resolve this, which requires depth measurements or an additional depth estimation model (and depth errors can degrade interpolation quality).

More specifically, we derive an occlusion-aware weighting mask to be used in warping. 
First, we obtain occlusion map $O_{0, 1}$ via forward-backward consistency \cite{meister2018unflow}, which indicates occlusion region on $I_0$; see how to compute forward-backward consistency in Eq. 1 of~\cite{meister2018unflow}. 
Next, we apply backward warping to $O_{0,1}$ using $V_{0\rightarrow1}$ and by further applying the non-occlusion mask $1-O_{0,1}$, 
% : $(1-O_{0,1})\cdot \omega_{b}(O_{0,1}, V_{0\rightarrow1})$, 
we can infer the pixels in $I_0$ that produce occlusion, i.e., foreground pixels, and accordingly generate the mask to select these pixels. More specifically, the occlusion-aware weighting mask is computed as follows: \vspace{-6pt}
\begin{equation} \label{weighting_masking}
\begin{split}
M_{0} = \alpha \cdot (1-O_{0,1})\cdot \omega_{b}(O_{0,1},\, V_{0\rightarrow1}),\\[-6pt]
\end{split}
\end{equation}
where $\alpha$ is a coefficient controlling the weighting.
% (We use $\alpha$ = 50) 

Fig.~\ref{fig:weighting_mask} provides a visual example. Assume that the disk moves to the right from $I_0$ to $I_1$ and the background is static. As indicated by $O_{0,1}$, the disk is not occluded and creates an occlusion blocking the background, i.e., it is the foreground. Due to this occlusion, when performing backward warping on $I_1$ or $O_{0,1}$ using $V_{0\rightarrow1}$, we not only move the disk back to its original location at $t=0$, but also create a duplicate at its current location since $V_{0\rightarrow1}$ is zero for the background (known as the ghosting effect~\cite{zhao2020maskflownet}). By applying the non-occlusion mask, we can remove the ghosting effect and recover the disk's original location; in other words, we obtain the weighting mask for selecting foreground pixels. 

We replace the depth-based masking in Eq.~\ref{real_interframe3} with our occlusion-aware weighting mask $M_{0}$ and apply forward warping in Eq.~\ref{forward_warp}, which gives \vspace{-2pt}
\begin{equation} \label{our_weighting_mask_interframe}
\widehat{V}_{t\rightarrow 1}(p) = 
\frac{\Sigma_{q} \text{exp}(M_{0}(q)) \cdot (1-t)\cdot V_{0\rightarrow 1} \cdot b(u)}{\Sigma_{q} \text{exp}(M_{0}(q)) \cdot b(u)},\\[-4pt]
% I_{t\rightarrow 1}(p) = 
% \frac{\Sigma_{q} \text{exp}(M_{0}(q)) \cdot I_{0}(q) \cdot b(u)}{\Sigma_{q} \text{exp}(M_{0}(q)) \cdot b(u)},
\end{equation}
where $q$, $p$, and $b(u)$ are from Eq.~\ref{real_interframe}. $V_{t\rightarrow 0}$ can be computed in a similar way.
% and $I_{1\rightarrow t}$ can be computed similarly.

\vspace{3pt}
\noindent \textbf{Hole Filling Using Optical Flow Consistency}
% \subsection{Hole Filling for Warped Optical Flow Maps}

Assuming linear motion, the directions in the optical flows $V_{t \rightarrow 0}$ and $V_{t \rightarrow 1}$ should be exactly opposite, and the magnitude ratio of the two flows should be $t:(1-t)$. Based on this assumption, we fill in the missing values in the synthesized optical flow maps as follows: 
\begin{equation} \label{fill_v1}
V_{t \rightarrow 0}(p) =
-\frac{t}{1-t}V_{t \rightarrow 1}(p),\quad \text{if} \ \Sigma_{q} b(u) = 0,
\end{equation}
% \begin{equation} \label{fill_v1}
% V_{t \rightarrow 0}(p) = \begin{cases}
% -\frac{t}{1-t}V_{t \rightarrow 1}(p), & \mbox{if} \ \Sigma_{q} b(u) = 0 \\
% V_{t \rightarrow 0}(p), & \mbox{otherwise}.
% \end{cases} 
% \end{equation}
where $\Sigma_{q} b(u) = 0$ indicates that no source pixels are assigned to the target location, thus creating a hole.

Now that we can resolve the ambiguities and missing values in forward warping, we can use our computed $\widehat{V}_{t\rightarrow0}$ and $\widehat{V}_{t\rightarrow1}$, along with the confidence maps, to perform backward warping and fusion to produce the inter-frame, based on Eq.~\ref{combine}.

% \begin{equation} \label{fill_v2}
% V_{t \rightarrow 1}(p) = \begin{cases}
% -\frac{1-t}{t}V_{t \rightarrow 0}(p), & \mbox{if} \ \Sigma_{q} b(u) = 0 \\
% V_{t \rightarrow 0}(p), & \mbox{otherwise}.
% \end{cases} 
% \end{equation}

\subsection{Teacher-Student Semi-Supervised Learning}
\label{sec:method_self}
\vspace{-3pt}

The generated intermediate video frames and optical flows create an opportunity to significantly augment the training of optical flow models. We propose a new semi-supervised training strategy to leverage the new image pairs and flows.

More specifically, we adopt a teacher-student training approach. The teacher network consumes an original pair of video frames and predicts the forward and backward optical flows. By using our proposed video interpolation algorithm, we can generate an inter-frame $I_t$ as well as the corresponding inter-flows $V_{0\rightarrow t}$ and $V_{t\rightarrow 0}$, for any intermediate time step. We randomly sample $t\in [0,1]$ to expand motion diversity in training data. We use $(I_0, I_t)$ to form a new training pair. The generated intermediate optical flows not only supply the supervision signal for the new pair, but also allows us to compute a confidence mask to make training more stable, i.e., we only train the network with the reliable optical flows. We train the student model with these new pairs derived from interpolation as well as with the available pseudo ground-truth data. Inspired by Mean Teacher~\cite{tarvainen2017mean}, we employ the Exponential Moving Average (EMA) update method to update the teacher model with a temporal ensemble of the student model.  

In addition, as inspired by Smurf~\cite{stone2021smurf}, RealFlow~\cite{han2022realflow}, and DistractFlow~\cite{jeong2023distractflow}, we further impose semantic distraction to $I_{0}$ to generate an augmented version $D_{\lambda}(I_{0}, \tilde{I}_d)$ (see DistractFlow for details of the frame distraction operation), and crop $I_{t}$ and $D_{\lambda}(I_{0}, \tilde{I}_d))$. 
Our self-supervision loss is given as follows for each new pair:\vspace{-5pt}
\begin{equation} \label{eq:semi_1}
\mathcal{L}_\text{self} =[C_{t,\,0} \geq \tau]\| f_\text{student} (I_{t}, D_{\lambda}(I_{0},\, \tilde{I}_d)) - \widehat{V}_{t \rightarrow 0}\|_{1}, \\
\end{equation}
where $f_\text{student} (I_{t}, D_{\lambda}(I_{0}, \tilde{I}_d))$ denotes the optical flow prediction on frame pair $I_t$ and $D_{\lambda}(I_{0}, \tilde{I}_d))$ by the student network, $C_{t,0}$ is the confidence map derived from $\widehat{V}_{t\rightarrow 0}$ and $\widehat{V}_{0\rightarrow t}$ based on Eq.~\ref{Con_map}, and $\tau$ is a threshold.

Combining this with the supervision from available ground-truth data, the total training loss for the student network is given by \vspace{-6pt}
\begin{equation} \label{eq:total_semi_loss}
\mathcal{L}_\text{total} = \mathcal{L}_\text{sup} + w \cdot \mathcal{L}_\text{self},\\[-5pt]
% \mathcal{L}_\text{total} = \mathcal{L}_\text{base} + w_\text{dist} \mathcal{L}_\text{dist} + w_\text{self} \mathcal{L}_\text{self},
\end{equation}
where $w$ is a weighting coefficient.

\begin{table*}[h]
\begin{center}
\caption{Video Frame Interpolation (VFI) results on Sintel (Clean), KITTI datasets. There are five state of the art backward warping based VFI algorithms (First to fifth rows), two forward warping based VFI algorithms (sixth to seventh) and ours (bottom). Algorithms of first to sixth rows trained their model on Vimeo dataset using pre-trained optical flow model, and two algorithms of seventh and eighth use RAFT trained on FlyingChairs and FlyingThings3D. A/V denotes AlexNet/VGG used in LPIPS. 
% \textcolor{red}{\textbf{Red}}/\textcolor{blue}{\textbf{Blue}}: Best and second best results.
\textbf{Bold}/\underline{Underline}: Best and second best results.
}
\vspace{-3mm}
\label{tab:vfi}
% \adjustbox{max width=0.48\textwidth}
\adjustbox{max width=1.0\textwidth}
{
\begin{tabular}{|l||cc|cc|cc||c|}
\hline
\multirow{2}{*}{Method} & \multicolumn{2}{|c|}{Sintel (12FPS $\rightarrow$ 24 FPS)} & \multicolumn{2}{|c|}{Sintel (6FPS $\rightarrow$ 12 FPS)} & \multicolumn{2}{|c|}{KITTI (5FPS $\rightarrow$ 10 FPS)} & Param  \\
\cline{2-7}
& PSNR / SSIM $\uparrow$ & LPIPS (A) / (V) $\downarrow$ & PSNR / SSIM $\uparrow$ & LPIPS (A) / (V) $\downarrow$ & PSNR / SSIM $\uparrow$ & LPIPS (A) / (V) $\downarrow$ & (M) \\
\hline
% \hline
IFRNet-B \cite{kong2022ifrnet} (CVPR 2022)  & 30.06 / 0.901 & 0.093 / 0.128 & 25.87 / 0.837 & 0.154 / 0.195& 21.64 / 0.760 & 0.140 / 0.217 & 5.0  \\
VFIFormer \cite{lu2022video} (CVPR 2022)  & 30.37 / \textbf{0.909} & 0.088 / \textbf{0.115} &  \underline{26.22} / \textbf{0.849} & 0.154 / 0.184& \textbf{22.53} / \textbf{0.787} & 0.149 / 0.227 & 24.1 \\
% VFIFormer \cite{lu2022video} (CVPR 2022)  & 30.37 / \textcolor{red}{\textbf{0.909}} & 0.088 / \textcolor{red}{\textbf{0.115}} &  \textcolor{blue}{\textbf{26.22}} / \textcolor{red}{\textbf{0.849}} & 0.154 / 0.184& \textcolor{red}{\textbf{22.53}} / \textcolor{red}{\textbf{0.787}} & 0.149 / 0.227 & 24.1 \\
RIFE \cite{huang2022real} (ECCV 2022) & 30.04 / 0.899 & 0.103 / 0.135 & 25.79 / 0.835 & 0.169 / 0.206 & 21.50 / 0.752 & 0.158 / 0.235 & 9.8 \\
EMA-VFI \cite{zhang2023extracting} (CVPR 2023) & \textbf{30.51} / 0.906 & 0.101 / 0.128 & \textbf{26.29} / 0.844 & 0.165 / 0.196 & 22.00 / 0.767 & 0.177 / 0.253 & 66.0 \\
% EMA-VFI \cite{zhang2023extracting} (CVPR 2023) & \textcolor{red}{\textbf{30.51}} / 0.906 & 0.101 / 0.128 & \textcolor{red}{\textbf{26.29}} / 0.844 & 0.165 / 0.196 & 22.00 / 0.767 & 0.177 / 0.253 & 66.0 \\
AMT-L \cite{li2023amt} (CVPR 2023) & 30.30 / 0.905 & 0.087 / 0.119 & 26.19 / 0.843 & 0.149 / 0.186 & 21.97 / 0.773 & 0.151 / 0.224 & 12.9 \\
\hline
SoftSplat \cite{niklaus2020softmax} (CVPR 2020) & \underline{30.42} / \underline{0.907} & 0.090 / 0.118 & 26.23 /  \underline{0.846} & 0.153 / 0.186 & 21.95 / \underline{0.776} & 0.169 / 0.244 & 12.2 \\
% SoftSplat \cite{niklaus2020softmax} (CVPR 2020) & \textcolor{blue}{\textbf{30.42}} / \textcolor{blue}{\textbf{0.907}} & 0.090 / 0.118 & 26.23 /  \textcolor{blue}{\textbf{0.846}} & 0.153 / 0.186 & 21.95 / \textcolor{blue}{\textbf{0.776}} & 0.169 / 0.244 & 12.2 \\
RIPR \cite{han2022realflow} (ECCV 2022) & 28.26 / 0.894 & \underline{0.084} / 0.120 & 24.91 / 0.826  & \underline{0.135} / \underline{0.181} & 21.18 / 0.733 & \textbf{0.112} / \underline{0.195} & 349.5  \\
% RealFlow \cite{han2022realflow} (ECCV 2022) & 28.26 / 0.894 & \textcolor{blue}{\textbf{0.084}} / 0.120 & 24.91 / 0.826  & \textcolor{blue}{\textbf{0.135}} / \textcolor{blue}{\textbf{0.181}} & 21.18 / 0.733 & \textcolor{red}{\textbf{0.112}} / \textcolor{blue}{\textbf{0.195}} & 349.5  \\
\rowcolor{lightgray} OCAI (ours) & 29.47 / 0.904 & \textbf{0.078} / \underline{0.117} & 25.88 / 0.838 & \textbf{0.129} / \textbf{0.178} & \underline{22.08} / 0.758 & \textbf{0.112} / \textbf{0.190} & 5.3 \\
% \rowcolor{lightgray} MoreFlow (Our) & 29.47 / 0.904 & \textcolor{red}{\textbf{0.078}} / \textcolor{blue}{\textbf{0.117}} & 25.88 / 0.838 & \textcolor{red}{\textbf{0.129}} / \textcolor{red}{\textbf{0.178}} & \textcolor{blue}{\textbf{22.08}} / 0.758 & \textcolor{red}{\textbf{0.112}} / \textcolor{red}{\textbf{0.190}} & 5.3 \\
\hline
\end{tabular}
}
\vspace{-7mm}
\end{center}
\end{table*}

\vspace{2pt}
\section{Experiment}
\label{sec:exp}
\vspace{-3pt}
We evaluate our proposed \ours method for video interpolation and for semi-supervised optical flow training on benchmark datasets, and compare \ours with baselines and latest state-of-the-art (SOTA) methods.

\begin{figure}[t!]
% \begin{figure}[htbp]
\begin{center}$
\centering
\begin{tabular}{c}
\hspace{-0.3cm} \includegraphics[width=8.3cm]{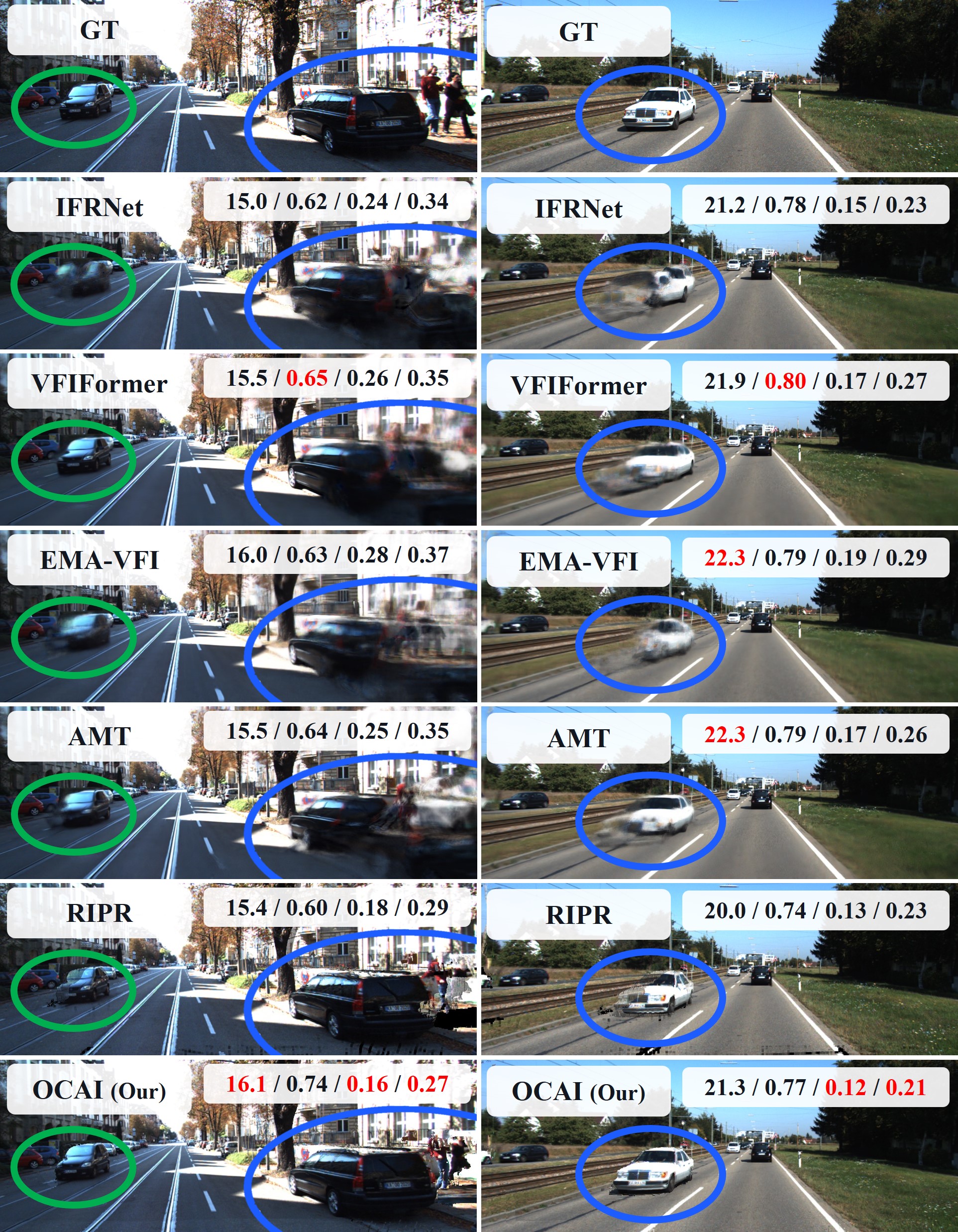}\\

% \hspace{-0.3cm} \includegraphics[width=4.1cm]{img/results/GT_007_v4.jpg}
% & \hspace{-0.33cm}\includegraphics[width=4.1cm]{img/results/GT_675_v4.jpg} \\

% \hspace{-0.3cm}  \includegraphics[width=4.1cm]{img/results/IFR_007_v4.jpg}
% & \hspace{-0.4cm} \includegraphics[width=4.1cm]{img/results/IFR_675_v4.jpg}\\

% \hspace{-0.3cm} \includegraphics[width=4.1cm]{img/results/VFIF_007_v4.jpg}
% & \hspace{-0.45cm} \includegraphics[width=4.15cm]{img/results/VFIF_675_v4.jpg}\\

% \hspace{-0.3cm} \includegraphics[width=4.1cm]{img/results/Ema_007_v4.jpg}
% & \hspace{-0.4cm} \includegraphics[width=4.1cm]{img/results/Ema_675_v4.jpg}\\

% \hspace{-0.3cm} \includegraphics[width=4.1cm]{img/results/AMT007_v4.jpg}
% & \hspace{-0.4cm} \includegraphics[width=4.1cm]{img/results/AMT675_v4.jpg}\\

% \hspace{-0.3cm} \includegraphics[width=4.1cm]{img/results/Real_007_v4.jpg}
% & \hspace{-0.4cm} \includegraphics[width=4.1cm]{img/results/Real_675_v4.jpg}\\

% \hspace{-0.3cm} \includegraphics[width=4.1cm]{img/results/Our_007_v4.jpg}
% & \hspace{-0.4cm} \includegraphics[width=4.1cm]{img/results/Our_675_v4.jpg}

\end{tabular}$
\end{center}
\vspace{-20pt}
\caption{\small Video Frame Interpolation (VFI) results on KITTI. First row is the ground truth. Second to fifth rows are outputs of SOTA VFI models~\cite{kong2022ifrnet, lu2022video, zhang2023extracting, li2023amt}. Sixth row is the output of using RealFlow~\cite{han2022realflow} for VFI. Bottom row shows our OCAI results. 
For each interpolated frame, we show the PSNR, SSIM, and LPIPS (using AlexNet and VGG) scores. Best scores are shown in red.\protect\footnotemark}
\label{exp:qualitative_vfi}
\vspace{-10pt}
% \vspace{-5pt}
\end{figure}
\footnotetext{Note that PSNR and SSIM do not always correlate well with perceived visual quality, as shown in previous studies~\cite{zhang2018unreasonable, prashnani2018pieapp}. For instance, in Fig.~\ref{exp:qualitative_vfi}, VFIFormer has higher SSIM scores but worse visual quality.}

\subsection{Experimental Setup}
\vspace{-3pt}
\textbf{Video Frame Interpolation (VFI): }
We compare with latest SOTA VFI algorithms~\cite{kong2022ifrnet, lu2022video, huang2022real, zhang2023extracting, li2023amt}. We use their official codes and weights trained on Vimeo90k~\cite{xue2019video}. In addition, we compare with RIPR of RealFlow~\cite{han2022realflow} as our forward warping baseline. We use their official code to generate inter-frames and RAFT~\cite{teed2020raft} trained on FlyingChair (C)~\cite{dosovitskiy2015flownet} and FlyingThings3D (T)~\cite{mayer2016large} as the optical flow model. 
In OCAI, we use the same optical flow network, i.e., RAFT trained on C+T, for fair comparison. 
% \js{We set $\alpha$ to 50 in Eq.~\ref{weighting_masking}}.
For evaluation, we use Sintel (S)~\cite{butler2012naturalistic} and KITTI (K)~\cite{geiger2013vision, menze2015object}, which are standard optical flow datasets.\footnote{The frames per second (FPS) numbers of Sintel and KITTI are 24 and 12, respectively.} More specifically, each test sample consists of three consecutive frames, with the first and third used as existing frames, and the second as the interpolation target. We use common image similarity metrics to evaluate VFI quality, including PSNR, SSIM~\cite{wang2004image}, and LPIPS (using AlexNet and VGG)~\cite{zhang2018unreasonable}. More details can be found in the supplementary file.

\noindent \textbf{Semi-Supervised Optical Flow (SSOF): }
We use RAFT as the network architecture, following previous semi-supervised optical flow model training settings~\cite{han2022realflow, im2022semi, jeong2023distractflow}. When evaluating on Sintel (train) and SlowFlow~\cite{Janai2017SlowFlow}, we first pretrain the network on C+T and then, use FlyingThings3D (T) as the labeled dataset and Sintel (S) as unlabeled dataset. For KITTI (train) evaluation, we use FlyingThings3D as the labeled dataset and KITTI (multiview) test as the unlabeled dataset, with initialization from C+T pre-trained weight. For Sintel and KITTI (test) evaluations, we use the same labeled datasets (i.e., C+T+S+K+HD1K~\cite{kondermann2016hci}) following the original RAFT supervised training setting, and use Sintel training (($I_t$, $I_{t+2}$) pairs), Monkaa~\cite{mayer2016large}, and KITTI (multiview) training dataset as unlabeled data. Note that Sintel and KITTI test sets are not used as unlabeled data for training in these test evaluations. 

\begin{table*}[t]
\begin{center}
\caption{Optical flow results on SlowFlow, Sintel (train), and KITTI (train) datasets. We train the model on FlyingChairs (C) and FlyingThings3D (T) as labeled data, and Sintel and KITTI multiview (S/K) as unlabeled data. BD in SlowFlow represents Blur Duration. \textbf{Bold}/\underline{Underline}: Best and second best results.
% \jl{How about we follow general conventions such as by using bold fonts and underlines or superscript marks to indicate best and second best entries?}
}
\vspace{-3mm}
\label{tab:ssof}
\adjustbox{max width=1.0\textwidth}
{
\begin{tabular}{|l||c|c||c|c|c|c|c|c|}
\hline
% Method & Labeled & Network & \multicolumn{2}{|c|}{mAP (\%)}\\
\multirow{2}{*}{Method} &  Labeled & Unlabeled    & \multicolumn{2}{|c|}{SlowFlow (100px)} & \multicolumn{2}{|c|}{Sintel (train)}  & \multicolumn{2}{|c|}{KITTI (train)}\\
\cline{4-9}
 & data & data & (3BD/epe) & (5BD/epe)& (Clean-epe) & (Final-epe) & (Fl-epe) & (Fl-all) \\

\hline
\hline
RAFT-Supervised &  C + T &  & 7.98 & 6.72& 1.43 &  2.71 & 5.04 & 17.4 \\
% \cline{4-5}
\hline
% \cline{2-2}
RAFT-A \cite{sun2021autoflow} (CVPR 2021) & AutoFlow \cite{sun2021autoflow}  &  & - & - & 1.95  & 2.57 &  4.23 & - \\
\cline{2-3}
RAFT-OCTC \cite{jeong2022imposing} (CVPR 2022) & \multirow{6}{*}{C + T} & T (subsampled)  & - & - & 1.31 & 2.67 & 4.72 & 16.3  \\
\cline{3-3}
Fixed Teacher \cite{im2022semi} (ECCV 2022) &   & \multirow{5}{*}{$S$/$K$} & - & - & 1.32 & 2.58 & 4.91 & 15.9\\
FlowSupervisor \cite{im2022semi} (ECCV 2022) & & &- & - & 1.30 &  2.46 & 3.35 & 11.1  \\
RealFlow \cite{han2022realflow} (ECCV 2022) &  &  & - & - & 1.34 &  2.38 & \underline{2.16} & \underline{8.5}  \\
DistractFlow \cite{jeong2023distractflow} (CVPR 2023) & &  & \underline{3.60} & \underline{5.15} & \underline{1.25} & \underline{2.35} & 3.01 & 11.7\\
% DistractFlow \cite{jeong2023distractflow} (CVPR 2023) & &  & \textcolor{blue}{\textbf{3.60}} & \textcolor{blue}{\textbf{5.15}} & \textcolor{blue}{\textbf{1.25}} & \textcolor{blue}{\textbf{2.35}} & 3.01 & 11.7\\
\rowcolor{lightgray} \ours (ours) & & & \textbf{2.97} & \textbf{5.04} & \textbf{1.20} & \textbf{2.32} & \textbf{2.07} & \textbf{7.6} \\
% \rowcolor{lightgray} MoreFlow (Our) & & & \textcolor{red}{\textbf{2.97}} & \textcolor{red}{\textbf{5.04}} & \textcolor{red}{\textbf{1.20}} & \textcolor{red}{\textbf{2.32}} & \textcolor{red}{\textbf{2.07}} & \textcolor{red}{\textbf{7.6}} \\
\hline
\end{tabular}
}
\vspace{-9mm}
\end{center}
\end{table*}

\begin{table}[t]
\begin{center}
\caption{Optical flow results on Sintel and KITTI test. * indicates ``warm-start" results that use previous flow prediction.
% RAFT-supervised model has been trained on C+T+S+K+H datasets, RAFT-A model trained on A (AutoFlow) +T+S+K+H, and FlowSupervisor trained on C+T+S+K+H as labeled dataset and S/K/Spring datasets as unlabeled dataset. RealFlow
}
\vspace{-3mm}
\label{tab:ssof_test}
\adjustbox{max width=0.45\textwidth}
{
\begin{tabular}{|l||c|c|}
\hline
% Method & Labeled & Network & \multicolumn{2}{|c|}{mAP (\%)}\\
\multirow{2}{*}{Method}  & Sintel (test)  & KITTI (test) \\
% \cline{2-4}
 & (Final-epe) & (Fl-all)\\

\hline
\hline
RAFT-Supervised  &3.18/2.86* & 5.10 \\
\hline
RAFT-A \cite{sun2021autoflow} (CVPR 2021) &3.14 &4.78 \\
RAFT-OCTC \cite{jeong2022imposing} (CVPR 2022)  &3.09 &4.72 \\
FlowSupervisor \cite{im2022semi} (ECCV 2022)  &2.79* &4.85 \\
RealFlow \cite{han2022realflow} (ECCV 2022)&- &4.63 \\
DistractFlow \cite{jeong2023distractflow} (CVPR 2023) &2.71* &4.71 \\
\rowcolor{lightgray} \ours (ours)  &\textbf{2.63}* &\textbf{4.13} \\

% \hline
% \hline
% RAFT-Supervised  & 1.94/1.61* & 3.18/2.86* & 5.10 \\
% \hline
% RAFT-A \cite{sun2021autoflow} (CVPR 2021) &  2.01 & 3.14 & 4.78 \\
% RAFT-OCTC \cite{jeong2022imposing} (CVPR 2022) &1.82 & 3.09 & 4.72 \\
% FlowSupervisor \cite{im2022semi} (ECCV 2022) & 1.65* & 2.79* & 4.85 \\
% RealFlow \cite{han2022realflow} (ECCV 2022)& -  & -   & 4.63 \\
% DistractFlow \cite{jeong2023distractflow} (CVPR 2023) & -  & 2.71* & 4.71 \\
% \rowcolor{lightgray} \ours (ours)  & & & \textbf{4.13} \\

\hline
\end{tabular}
}
\vspace{-5mm}
\end{center}
\end{table}

\subsection{Video Frame Interpolation} \vspace{-3pt}
Table \ref{tab:vfi} and Fig.~\ref{exp:qualitative_vfi} show the performance of SOTA VFI methods, RIPR of RealFlow, and our method on Sintel (clean) and KITTI. 
We achieve the best LPIPS scores and PSNR/SSIM scores on-par with existing SOTA solutions; note that PSNR and SSIM do not always correctly reflect visual quality~\cite{prashnani2018pieapp}.
Since SOTA methods use backward warping and predict intermediate two optical flows ($V_{t \rightarrow 0}$ and $V_{t \rightarrow 1}$) without the inter-frame $I_{t}$, they do not work well when there are large displacements, which results in blurriness (see cars in Fig~\ref{exp:qualitative_vfi}). In contrast, forward warping methods (RIPR of RealFlow and ours) can predict accurate $V_{0 \rightarrow 1}$ and $V_{1 \rightarrow 0}$, and handles fast moving objects better. Furthermore, our OCAI method produces better VFI quality than RealFlow, with sharper image details and fewer holes. 
More interpolation results using different $t$ values  (e.g., 0.2, 0.4, 0.6, 0.8) can be found in the supplementary material.

% We further provide inter-frames using our method for 051
% different t values (e.g., 0.2, 0.4, 0.6, 0.8) in the supp.
% While backward warping based state of the art VFI methods show high PSNR and SSIM scores on Sintel and KITTI datasets, RealFlow and our model shows lower LPIPS scores on both datasets. Since backward warping based algorithms predict the two optical flows ($V_{t \rightarrow 0}$ and $V_{t \rightarrow 1}$) without inter-frame $I_{t}$, it cannot predicts large displacement and makes it blur in large motion region (Cars in Fig~\ref{exp:qualitative_vfi}). On the other hand, forward warping methods (RealFlow and OCAI) can predict accurate two optical flows ($V_{0 \rightarrow 1}$ and $V_{1 \rightarrow 0}$) and utilize them to generate large displacement objects well.

OCAI only requires the RAFT optical flow model with 5.3M parameters and shows better VFI performance as compared to existing SOTA methods using similar or more network parameters. Backward-warping-based approaches require additional training on Vimeo data with a pre-trained optical flow model, while RIPR of RealFlow and our OCAI only needs the pre-trained optical flow model without any additional training. Moreover, we achieve better VFI scores than RIPR \ignore{RealFlow} on both Sintel and KITTI. These results demonstrate that OCAI generates accurate inter-frames without needing depth estimation in forward warping. 

% Our Moreflow requires 5.3M parameters, and it shows better performance except for Sintel PSNR compared to IFRNet with similar parameters. Note that backward warping based algorithms are additionally trained on Vimeo dataset using pre-trained optical flow model, while RealFlow and our model use only pre-trained optical flow model without additional training. In addition, Our MoreFlow shows higher PSNR and SSIM scores and lower LPIPS scores than RealFlow on Sintel and KITTI without additional depth information. These results show our OCAI can be replaced by depth estimation model in forward warping process and generate accurate inter-frame. 

\subsection{Optical Flow} \vspace{-3pt}
Table~\ref{tab:ssof} shows the optical flow estimation evaluation results on SlowFlow, Sintel, and KITTI. We see that our proposed OCAI achieves the best performance on all datasets. Notably, it has significantly more accurate optical flow estimation as compared to latest SOTA such as RealFlow and DistractFlow. Specifically, on KITTI, OCAI brings nearly 1-point reduction in Fl-epe when comparing to DistractFlow and 1-point smaller Fl-all than RealFlow.

% RealFlow shows the second best performance on KITTI dataset and DistractFlow shows the second best performance on Sintel dataset. Our MoreFlow, which takes advantages of RealFlow and DistractFlow, achieves the best performance on all datasets. 

Table~\ref{tab:ssof_test} shows the evaluation result on KITTI test dataset. We achieve SOTA performance in semi-supervised optical flow on KITTI, bringing a significant improvement as compared to existing SOTA semi-supervised optical flow algorithms.  

% \hc{Keep or remove Sintel (test) depending on whether we could get Sintel result from server.} 
% \js{Sintel homepage is still down. We can access to the homepage, but we cannot upload our results. (If I couldn't get the number by noon, I will remove Sintel test results.)}
% Table~\ref{tab:ssof_test} shows the optical flow results of Sintel and KITTI test datasets. Our MoreFlow shows the state of the art performance on KITTI dataset in Semi-Supervised Optical Flow. Especially, our MoreFlow shows large gap between previous Semi-supervised optical flow algorithms on KITTI test dataset. 

\section{Ablation Studies}
% \section{Ablation Study and Discussion}
% I think discussion includes ablation study, so I don't think we need to mention both.  
\label{sec:dis}
\vspace{-3pt}

\begin{table}[t]
\begin{center}
\caption{Ablation Study of video frame interpolation on KITTI. $M$ refers to using our occlusion-aware weighting mask (Eq.~\ref{weighting_masking}) in $Z$ (Eq.~\ref{real_interframe3}). BHF is Bi-directional Hole Filling proposed in RealFlow, CBF refers to Confidence-Based Fusion (Eq.~\ref{combine}). 
% I is image warping and V is optical flow warping in $W$. $O$ is occlusion based Hole Filling ($H$) and Fusion ($F$) and $F$ (Eq.~\ref{fill_v1}) is optical flow consistency based Hole Filling ($H$). $C$ (Eq.~\ref{combine}) is confidence based Fusion ($F$). 
% \hc{How about we just show LPIPs (V)? We can use the extra space to expand Z W H F to more readable abbreviations perhaps, like Weighting, Warping, HoleFill, Fusion?} \js{Good idea. I am generating visual examples and also computing the Sintel scores. I will update them when I generate them.}
}
\vspace{-3mm}
\label{tab:vfi_ablation}
% \adjustbox{max width=0.48\textwidth}
\adjustbox{max width=0.48\textwidth}
{
\begin{tabular}{|l||cccc|cc|}
\hline
\multirow{2}{*}{Method} & \multirow{2}{*}{$Z$ (Eq.~\ref{real_interframe3})} & \multirow{2}{*}{Warping} & \multirow{2}{*}{Hole Filling} & \multirow{2}{*}{Fusion} & \multicolumn{2}{|c|}{KITTI}  \\
\cline{6-7}
& & & & & SSIM $\uparrow$ & LPIPS (V) $\downarrow$\\
\hline
\hline
% IFRNet-B \cite{kong2022ifrnet} (CVPR 2022)   & 21.64 & 0.760 & 0.140 & 0.217 \\
% AMT-L \cite{li2023amt} (CVPR 2023) & 21.97 & \textbf{0.773} & 0.151 & 0.224 \\
RIPR \cite{han2022realflow}  & Depth & Image & Image & BHF & 0.733 & 0.195 \\
\hline
\multirow{4}{*}{OCAI} & $M$ & Image & Image & BHF & 0.734 & 0.198 \\
 & $M$ & Flow & - & BHF & 0.721 & 0.213 \\ 
 & $M$ & Flow & Flow & BHF & 0.739 & 0.195 \\
 & $M$ & Flow & Flow & CBF & \textbf{0.758} & \textbf{0.190} \\
\hline
\end{tabular}
}
\vspace{-7mm}
\end{center}
\end{table}

\begin{figure}[t!]
% \begin{figure}[htbp]
\begin{center}$
\centering
\begin{tabular}{c}
\hspace{-0.3cm} \includegraphics[width=8.3cm]{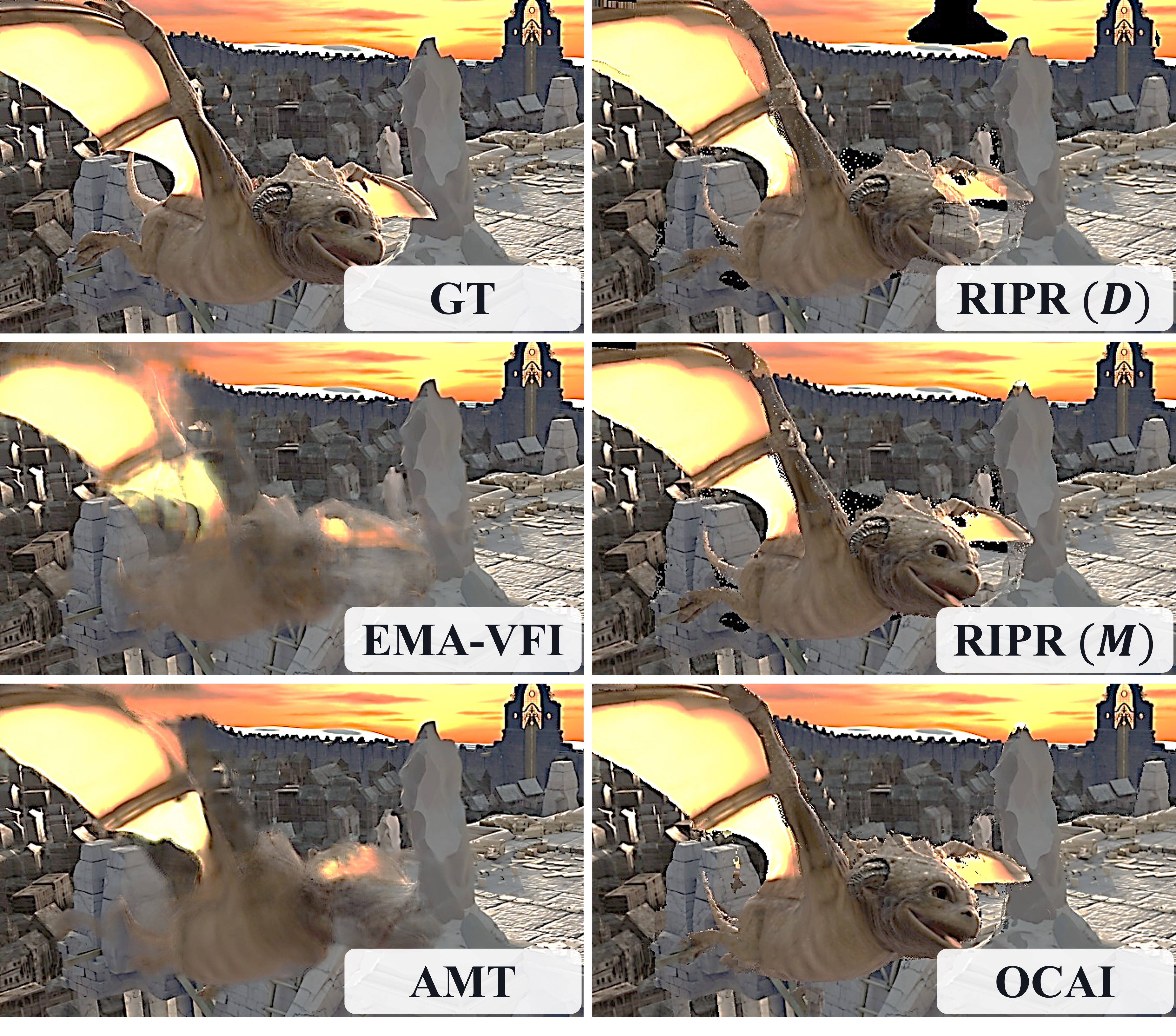}
% \hspace{-0.3cm} \includegraphics[width=2.7cm]{img/results/Sintel_gt3.jpg}
% & \hspace{-0.4cm} \includegraphics[width=2.7cm]{img/results/Sintel_ema.jpg}
% & \hspace{-0.4cm} \includegraphics[width=2.7cm]{img/results/Sintel_amt.jpg} \\
% \hspace{-0.3cm} \includegraphics[width=2.7cm]{img/results/Sintel_real_v3.jpg}
% & \hspace{-0.4cm} \includegraphics[width=2.7cm]{img/results/Sintel_m_v3.jpg}
% & \hspace{-0.4cm} \includegraphics[width=2.7cm]{img/results/Sintel_our_v3.jpg}\\
% \hspace{-0.3cm} \includegraphics[width=4.1cm]{img/results/Sintel_gt3.jpg}
% & \hspace{-0.3cm} \includegraphics[width=4.1cm]{img/results/Sintel_real_v3.jpg}  \\
% \hspace{-0.3cm} \includegraphics[width=4.1cm]{img/results/Sintel_ema.jpg}
% & \hspace{-0.3cm} \includegraphics[width=4.1cm]{img/results/Sintel_m_v3.jpg}\\
% \hspace{-0.3cm} \includegraphics[width=4.1cm]{img/results/Sintel_amt.jpg}
% & \hspace{-0.3cm} \includegraphics[width=4.1cm]{img/results/Sintel_our_v3.jpg}\\
\end{tabular}$
\end{center}
\vspace{-24pt}
\caption{\small Video Frame Interpolation (VFI) results on Sintel. RIPR (D) and RIPR (M) images are generated using RIPR with depth or occlusion weighting mask ($M$). 
% $OCAI^{*}$ and $OCAI^{\dagger}$ images are generated without hole filling, and $OCAI^{*}$ is applied BHF fusion. 
OCAI is generated using $M$, with flow hole-filling and confidence fusion. 
% \footnote{dd}
% \footnotemark{}
}
\label{exp:ablation_sintel_vfi}
% \vspace{-10pt}
\vspace{-8pt}
\end{figure}

\begin{figure}[t!]
% \begin{figure}[htbp]
\begin{center}$
\centering
\begin{tabular}{c}

\hspace{-0.3cm} \includegraphics[width=8.3cm]{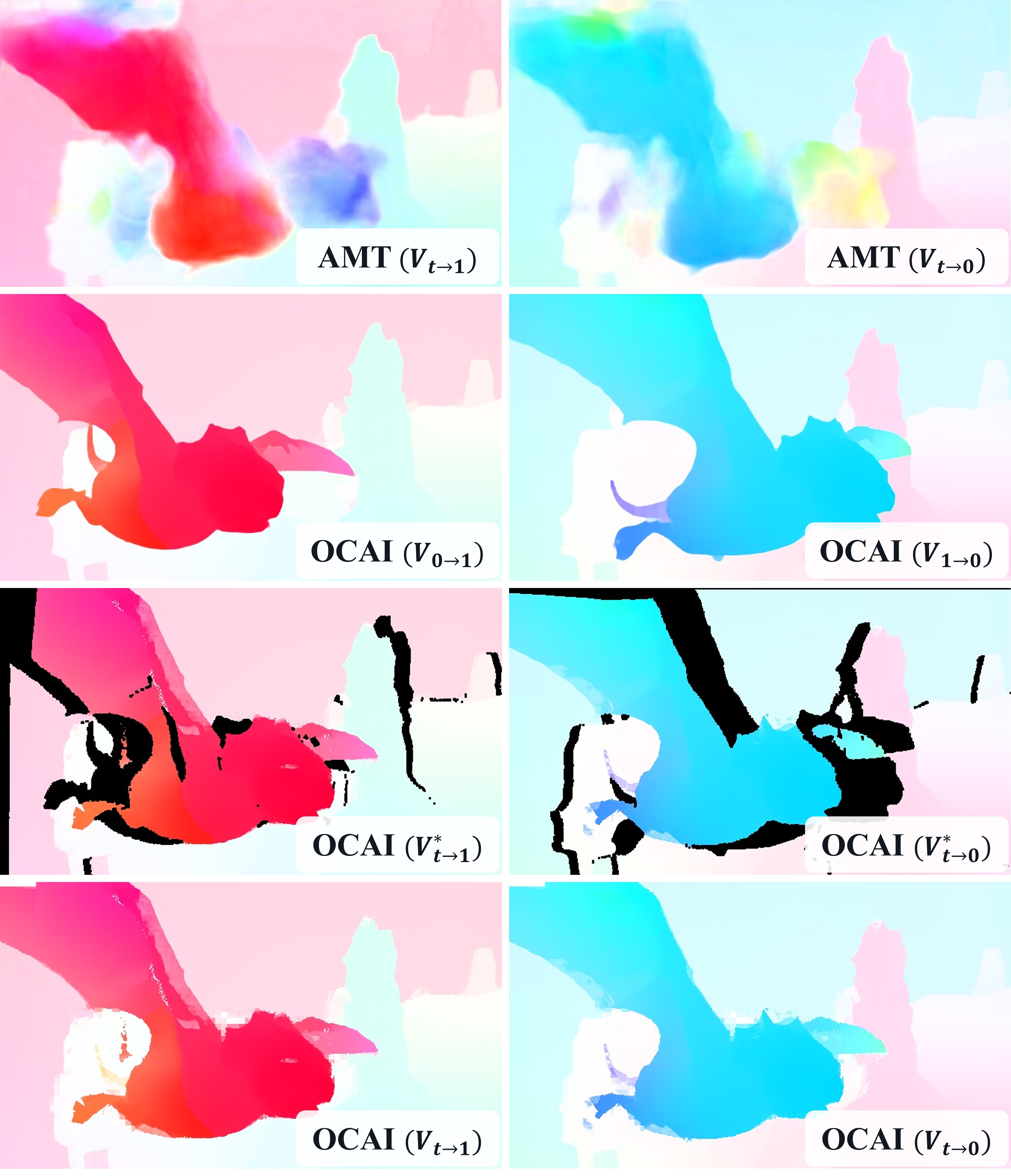}

% \hspace{-0.3cm} \includegraphics[width=4.1cm]{img/results/Sintel_flow_01.jpg}
% & \hspace{-0.3cm}\includegraphics[width=4.1cm]{img/results/Sintel_flow_10.jpg} \\

% % \hspace{-0.3cm}  \includegraphics[width=4.1cm]{img/results/sintel_check.jpg}
% \hspace{-0.3cm}  \includegraphics[width=4.1cm]{img/results/Sintel_flow_t1_wh.jpg}
% & \hspace{-0.4cm} \includegraphics[width=4.1cm]{img/results/Sintel_flow_t0_wh.jpg}\\

% \hspace{-0.3cm} \includegraphics[width=4.1cm]{img/results/Sintel_flow_t1_woh.jpg}
% & \hspace{-0.4cm} \includegraphics[width=4.1cm]{img/results/Sintel_flow_t0_woh.jpg}\\

% \hspace{-0.3cm} \includegraphics[width=4.1cm]{img/results/sintel_flow_amt_t1.jpg}
% & \hspace{-0.4cm} \includegraphics[width=4.1cm]{img/results/sintel_flow_amt_t0.jpg}\\

\end{tabular}$
\end{center}
\vspace{-20pt}
\caption{\small Forward warped Optical Flow on Sintel (Same images from \ref{exp:ablation_sintel_vfi}). In top, there are two optical flows ($V_{t \rightarrow 1}$, $V_{t \rightarrow 0}$) from SOTA VFI algorithm. Two optical flows from images are shown in second row, and forward warped flows with holes are shown in third row ($V^{*}_{t \rightarrow 1}$, $V^{*}_{t \rightarrow 0}$). There are hole filled optical flow in the bottom. (Since visualization code applies normalization, color intensity looks similar, but the magnitude was reduced and shift and hole filling are visible.)
% \js{keep only $V_{*\rightarrow 0}$ flows? (to reduce the space) In $V_{*\rightarrow 1}$, front of left wing has overlap between foreground and background, so, there is a color gap. (Nevertheless, our method works better than VFI method.)} \hc{Perhaps let's first finalize the other comments. We may have space to keep this figure in full.}
% \footnote{dd}
% \footnotemark{}
}
\label{exp:ablation_sintel_vfi_flow}
% \vspace{-10pt}
\vspace{-8pt}
\end{figure}

\subsection{Video Frame Interpolation}\vspace{-3pt}

\textbf{Depth weighting vs. occlusion-aware weighting. }
Table~\ref{tab:vfi_ablation} shows the effectiveness of our  occlusion- and consistency-Aware forward warping for VFI. The first row shows forward image warping using depth weighting. In the second row, we replace depth weighting with our occlusion-aware weighting mask, which shows comparable performance without using depth. Fig.~\ref{exp:ablation_sintel_vfi} provides sample qualitative results. RIPR (from RealFlow) using depth and using occlusion-aware weighting both have holes (See RIPR (D) and RIPR (M)). Since RIPR performs hole-filling using another warped image, a hole cannot be filled when both images have holes in the same corresponding regions (e.g., wings and head in Fig.~\ref{exp:ablation_sintel_vfi}).

% Table~\ref{tab:vfi_ablation} shows the effectiveness of our Occlusion- and Consistency-Aware forward warping based Interpolation (OCAI). In the top of Table~\ref{tab:vfi_ablation}, it shows the results of forward image warping using depth weighting. In the second row, we replace depth mask with occlusion aware weighting mask. Our proposed occlusion aware weighting mask shows comparable results compared to Depth weighting mask without depth information. Fig.~\ref{exp:ablation_sintel_vfi} shows some qualitative results. RIPR with depth in RealFlow and with occlusion weighting mask both still have holes (See RIPR (D) and RIPR (M) in Fig.~\ref{exp:ablation_sintel_vfi}). Since RIPR fills the hole with another warped image, hole still can be existed when both images have holes in the same region(wings and head in Fig.~\ref{exp:ablation_sintel_vfi}).

\vspace{1mm}
\textbf{Image warping vs. flow warping. }
In the third row of Table~\ref{tab:vfi_ablation}, we show the effect of applying forward warping to flow instead of image (see optical flows in the third row of Fig.~\ref{exp:ablation_sintel_vfi_flow}). 
After performing hole-filling based on optical flow consistency, our intermediate flows are significantly improved (see inpainted flows in the fourth row of Fig.~\ref{exp:ablation_sintel_vfi_flow}).  
% In forward warping based fusion, because of fractional values in warping, there are some noise. 
By generating the confidence maps, we can then combine two warped images to more accurately generate the inter-frame (see OCAI output in Fig.~\ref{exp:ablation_sintel_vfi}).

\begin{table}[t]
\begin{center}
\caption{Ablation study of our semi-supervised optical flow training. We denote EMA, confidence-base loss masking, and image distraction as E, C, and D, respectively. We denote KITTI test dataset as K-Test, and RealFlow as RF.
% RealFlow and OCAI as RF and MF, respectively.
}
\vspace{-8pt}
\label{tab:ssof_ablation}
% \adjustbox{max width=0.48\textwidth}
\adjustbox{max width=0.48\textwidth}
{
\begin{tabular}{|l|c|c|c|c|c||cc|}
\hline
\multirow{2}{*}{Method} & \multirow{2}{*}{$E$} & \multirow{2}{*}{$C$} & \multirow{2}{*}{$D$} & \multirow{2}{*}{Dataset} & Training  & \multicolumn{2}{|c|}{KITTI 15} \\
& & & & & iterations  & (Fl-epe) & (Fl-all)\\ 
\hline
\hline
RAFT \cite{teed2020raft} & & & & &   & 5.04 & 17.4 \\
\hline
\multirow{2}{*}{RealFlow \cite{han2022realflow}} & & & & RF-K-Test & 200k (1 EM)  & 2.79 & 10.7 \\
 & & & & RF-K-Test & 800k (4 EM)  & 2.16 & 8.5 \\
\hline
DistractFlow & & & \checkmark & S/K-Test & 150k &3.01 & 11.7 \\
\hline
% \multirow{4}{*}{MoreFlow (Our)} 
&  & & & K-Test & 100k  & \multicolumn{2}{c|}{Diverged} \\
\multirow{2}{*}{\ours}  & \checkmark  & & & K-Test & 100k  & 2.85 & 10.1 \\
\multirow{2}{*}{(ours)} & \checkmark & \checkmark  & & K-Test & 100k  & 2.81 & 9.7 \\
& \checkmark  & \checkmark  & \checkmark  & K-Test & 100k  & 2.27 & 8.4 \\
% & \checkmark  & \checkmark  &  & IF-K-Test & 100k  & 2.62  & 8.9 \\
& \checkmark  & \checkmark  & \checkmark  & OCAI-K-Test & 100k & \textbf{2.07} & \textbf{7.6} \\
% Our & - & - & - & - \\
\hline
\end{tabular}
}
\vspace{-4mm}
\end{center}
\end{table}

\subsection{Semi-Supervised Optical Flow}
\vspace{-3pt}
\textbf{Using baseline VFI for semi-supervised learning.}
Table~\ref{tab:ssof_ablation} shows an ablation study of the semi-supervised training. RealFlow trains the model using EM algorithm, which requires significantly more iterations. 
% In addition, even with four times EM algorithms, teacher network is updated only four times. 
In our training, when only using EMA, our model already has a lower Fl-all score as compared to RealFlow with 1 EM iteration. By additionally masking the loss with confidence map and imposing image distractions, the model further improves. Note that EMA is crucial for training stability; the training fails to converge when teacher and student models are directly weight-shared. Finally, when using intermediate frames and flows generated by OCAI in training, our model achieves significantly lower Fl-epe and Fl-all. 
% \js{In addition, we evaluate VFI using our semi-supervised training weight which can generate more accurate inter-frame. The results of these experiments can be found in the supplementary material.}

% Table~\ref{tab:ssof_ablation} shows the ablation studies of our Semi-Supervised Optical Flow training. Since RealFlow trains the model in EM algorithm, it requires lots of training iterations. In addition, even with four times EM algorithms, teacher network is updated only four times. On the other hand, our algorithm uses EMA which can update teacher network using temporal ensenbled weights, and we can robustly train the network. In our training with only EMA method usint KITTI test dataset, our model shows less Fl-all score compared to 1 EM of RealFlow. It also shows additional improvement with computing only high confident area loss, and Distraction shows large improvement without inter-frame information. In addition, it shows further improvement using our OCAI based inter-frame and distraction in training process. 

% We train the network with 10 batch size (2 label (FlyingThings3D), 8 unlabel (K-Test)), and 100k iterations. 

\begin{table}[t]
\begin{center}
\caption{Effectiveness of new semi-supervised trained model for Video Frame interpolation. }
\vspace{-3mm}
\label{tab:ssof_vfi_help}
% \adjustbox{max width=0.48\textwidth}
\adjustbox{max width=0.48\textwidth}
{
\begin{tabular}{|c|c|c|cc|}
\hline
\multirow{2}{*}{VFI dataset} & \multicolumn{2}{|c|}{Flow Train Dataset} & \multirow{2}{*}{PSNR / SSIM $\uparrow$} & \multirow{2}{*}{LPIPS (A) / (V) $\downarrow$} \\
\cline{2-3}
& Label & Unlabel & &\\
\hline
Sintel & \multirow{6}{*}{C+T} & - & 29.47 / 0.904 & 0.078 / 0.117 \\
(12 FPS $\rightarrow$ 24 FPS)& & OCAI-S-Test & \textbf{29.51} / \textbf{0.905} & \textbf{0.756} / \textbf{0.115} \\
\cline{1-1}
\cline{3-5}
% \hline
Sintel &  & - & \textbf{25.88} / 0.838 & 0.129 / 0.178 \\
(6 FPS $\rightarrow$ 12 FPS)& & OCAI-S-Test & 25.87 / \textbf{0.840} & \textbf{0.127} / \textbf{0.176} \\
\cline{1-1}
\cline{3-5}
% \hline
KITTI &  & - & 22.08 / 0.758 & 0.112 / 0.190 \\
(5 FPS $\rightarrow$ 10 FPS)& & OCAI-K-Test & \textbf{22.29} / \textbf{0.759} & \textbf{0.108} / \textbf{0.186} \\
\hline
\end{tabular}
}
\vspace{-6mm}
\end{center}
\end{table}

\vspace{1mm}
\noindent \textbf{VFI using optical flow model trained from semi-supervised scheme.} 
% \hc{Do we need to show this in the main paper? Perhaps can move to supp to save space?} \js{Yes, we can just mention and upload in the supple.}
After we train the optical flow in semi-supervised manner, we evaluate the VFI performance of semi-supervised training in Table~\ref{tab:ssof_vfi_help}. Our semi-supervised training weight shows the improvements on all evaluation metrics on all dataset except for Sintel PSNR in 6 FPS to 12 FPS setting. While our \ours trains the optical flow in a semi-supervised manner, improved optical flow can generate more accurate inter-frame. Our \ours can boost the performances of not only optical flow but also VFI.

% Our MoreFlow trains the optical flow in a semi-supervised manner, and improved optical flow can be able to generate accurate optical flow. This can affect to the optical flow network training process. 

% \begin{table}[t]
% \begin{center}
% \caption{Effectiveness of new semi-supervised trained model for Video Frame interpolation. }
% % \vspace{-3mm}
% \label{tab:ssof_vfi_help}
% % \adjustbox{max width=0.48\textwidth}
% \adjustbox{max width=0.48\textwidth}
% {
% \begin{tabular}{|c|c|c|cccc|}
% \hline
% \multirow{2}{*}{VFI dataset} & \multicolumn{2}{|c|}{Flow Train Dataset} & \multirow{2}{*}{PSNR $\uparrow$} & \multirow{2}{*}{SSIM $\uparrow$} & \multirow{2}{*}{LPIPS (A) $\downarrow$} & \multirow{2}{*}{LPIPS (V) $\downarrow$} \\
% \cline{2-3}
% & Label & Unlabel & & & &\\
% \hline
% Sintel & \multirow{6}{*}{C+T} & - & 29.47 & 0.904 & 0.078 & 0.117 \\
% (12 FPS $\rightarrow$ 24 FPS)& & MF-S-Test & \textbf{29.51} & \textbf{0.905} & \textbf{0.756} & \textbf{0.115} \\
% \cline{1-1}
% \cline{3-7}
% % \hline
% Sintel &  & - & \textbf{25.88} & 0.838 & 0.129 & 0.178 \\
% (6 FPS $\rightarrow$ 12 FPS)& & MF-S-Test & 25.87 & \textbf{0.840} & \textbf{0.127} & \textbf{0.176} \\
% \cline{1-1}
% \cline{3-7}
% % \hline
% KITTI &  & - & 22.08 & 0.758 & 0.112 & 0.190 \\
% (5 FPS $\rightarrow$ 10 FPS)& & MF-K-Test & \textbf{22.29} & \textbf{0.759} & \textbf{0.108} & \textbf{0.186} \\
% \hline
% \end{tabular}
% }
% \vspace{-5mm}
% \end{center}
% \end{table}

\section{Conclusion}
\label{sec:conclusion}
\vspace{-3pt}
In this paper, we proposed a novel scheme that significantly augments the training of optical flow models. This effectively alleviates the lack of ground-truth optical flow labels in existing datasets. More specifically, we first proposed an occlusion-aware video frame interpolation method, which can robustly generate interframes despite large motions, as well as the intermediate optical flows. This allows us to significantly expand existing optical flow training data for free. We further proposed a semi-supervised training approach by leveraging the video frame interpolation. Through extensive experiments on standard optical flow benchmarks like Sintel and KITTI, we demonstrate the efficacy of our proposed approach and that it sets the new state of the art.

\newpage

{
    \small
    \bibliographystyle{ieeenat_fullname}
    \bibliography{main}
}

\newpage
\clearpage
\setcounter{page}{1}
\maketitlesupplementary

\section{Implementation and Training Details}
\label{supple:implementation}
\subsection{Video Frame Interpolation}

\textbf{Datasets.}
We use Sintel and KITTI datasets, which are standard Optical Flow datasets. Sintel (clean) dataset consists of 20 $\sim$ 50 consecutive frames in 23 Videos. In 12 FPS $\rightarrow$ 24 FPS frame interpolation, we load three consecutive frames ($I_{1}, I_{2}, I_{3}$), and use $I_{1}$ and $I_{3}$ as an input and generate $\hat{I}_{2}$ image. Then, we compute the PSNR, SSIM, and LPIPS (using AlexNet and VGG) metrics. And then, we load next frames ($I_{2}, I_{3}, I_{4}$), and generate $\hat{I}_{3}$ using $I_{2}$ and $I_{4}$ frames. We generate all frames from $\hat{I}_{2}$ to $\hat{I}_{N-1}$ images. Here, $N$ is the number of the frame in each video clip (total 1018 pairs). In 6 FPS $\rightarrow$ 12 FPS frame interpolation, we load ($I_{1}, I_{3}, I_{5}$), and generate $\hat{I}_{3}$. Then, we generate frames from $\hat{I}_{3}$ to $\hat{I}_{N-2}$ (total 972 pairs). KITTI (multiview) train dataset consists of 21 consecutive frames in 200 videos. We generate $\hat{I}_{2}$ to $\hat{I}_{20}$ frames, and there are 3800 pairs.

\textbf{Implementation Details. }
We use IFRNet \cite{kong2022ifrnet}, VFIFormer \cite{lu2022video}, RIFE \cite{huang2022real}, EMA-VFI \cite{zhang2023extracting}, and AMT \cite{li2023amt} VFI algorithms as our backward warping baselines. We use their official codes and weights trained on Vimeo90k.\footnote{IFRNet : \url{https://github.com/ltkong218/IFRNet}, VFIFormer: \url{https://github.com/dvlab-research/VFIformer}, RIFT: \url{https://github.com/megvii-research/ECCV2022-RIFE}, EMA-VFI: \url{https://github.com/MCG-NJU/EMA-VFI}, AMT: \url{https://github.com/MCG-NKU/AMT}}
We also use Soft-Splatting \cite{niklaus2020softmax} and RIPR of RealFlow \cite{han2022realflow} algorithm as our forward warping baselines. We use official codes, Vimeo trained weight for Soft-Splatting, and FlyingChairs+FlyingThings3D trained weight for RIPR.\footnote{Soft-Splatting: \url{https://github.com/sniklaus/softmax-splatting} RealFlow: \url{https://github.com/megvii-research/RealFlow}} RIPR and our OCAI use RAFT \cite{teed2020raft} optical flow model, and we also use the same weight with RIPR for fair comparison. We  set $\alpha$ in Eq.~10 to 50. Higher $\alpha$ shows good performance as shown in Table~\ref{supple_table1}. However, when it is set too high, e.g., above 100, the result becomes \textit{not a number}.

\begin{table}[h]
\begin{center}
\caption{Video Frame Interpolation results on KITTI. We evaluate the VFI with different $\alpha$ weights.}
\vspace{-3mm}
\label{supple_table1}
% \adjustbox{max width=0.48\textwidth}
\adjustbox{max width=0.48\textwidth}
{
\begin{tabular}{|c||cc|}
\hline
$\alpha$ & PSNR / SSIM $\uparrow$ & LPIPS (A) / (V) $\downarrow$ \\
\hline
1 &  21.98 / 0.756  &  0.114 /  0.192\\
10 & 22.06 / 0.757 & 0.112 / 0.191 \\
50 & \textbf{22.08} / \textbf{0.758} & \textbf{0.112} / \textbf{0.190} \\
100 & NA / NA & NA / NA \\
\hline
\end{tabular}
}
\vspace{-8mm}
\end{center}
\end{table}

\newpage

\begin{figure*}[p]
% \begin{figure}[htbp]
\begin{center}$
\centering
\begin{tabular}{c}
\hspace{-0.3cm} \includegraphics[width=17.4cm]{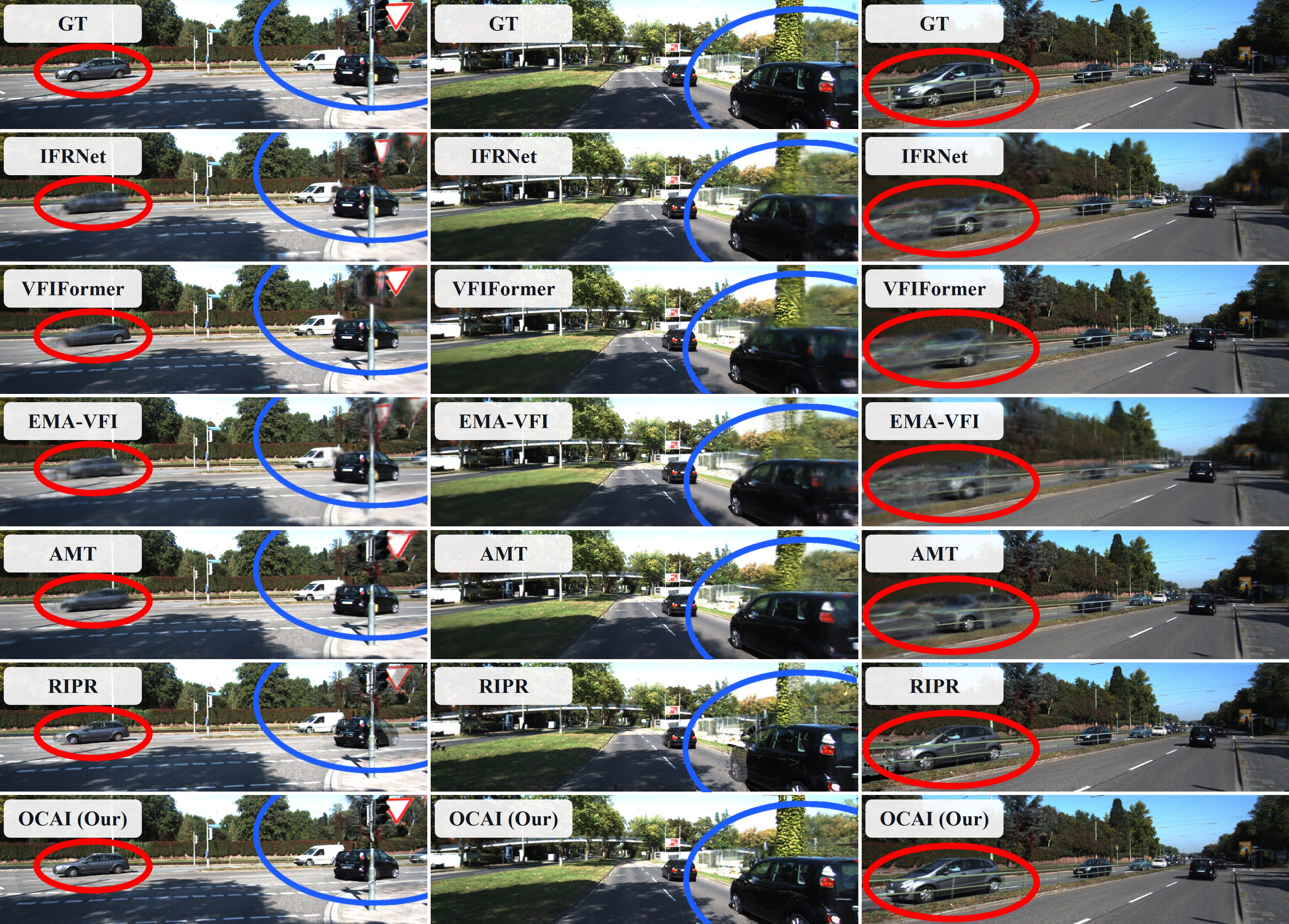}\\

\end{tabular}$
\end{center}
\vspace{-20pt}
\caption{\small Video Frame Interpolation (VFI) results on KITTI. First row is the ground truth. Second to fifth rows are outputs of SOTA VFI models~\cite{kong2022ifrnet, lu2022video, zhang2023extracting, li2023amt}. Sixth row is the output of using RealFlow~\cite{han2022realflow} for VFI. Bottom row shows our OCAI results. }
\label{exp:qualitative_vfi_kitti}
% \vspace{-10pt}
\vspace{-5pt}
\end{figure*}

\begin{figure*}[p]
% \begin{figure}[htbp]
\begin{center}$
\centering
\begin{tabular}{c}
\hspace{-0.3cm} \includegraphics[width=17.4cm]{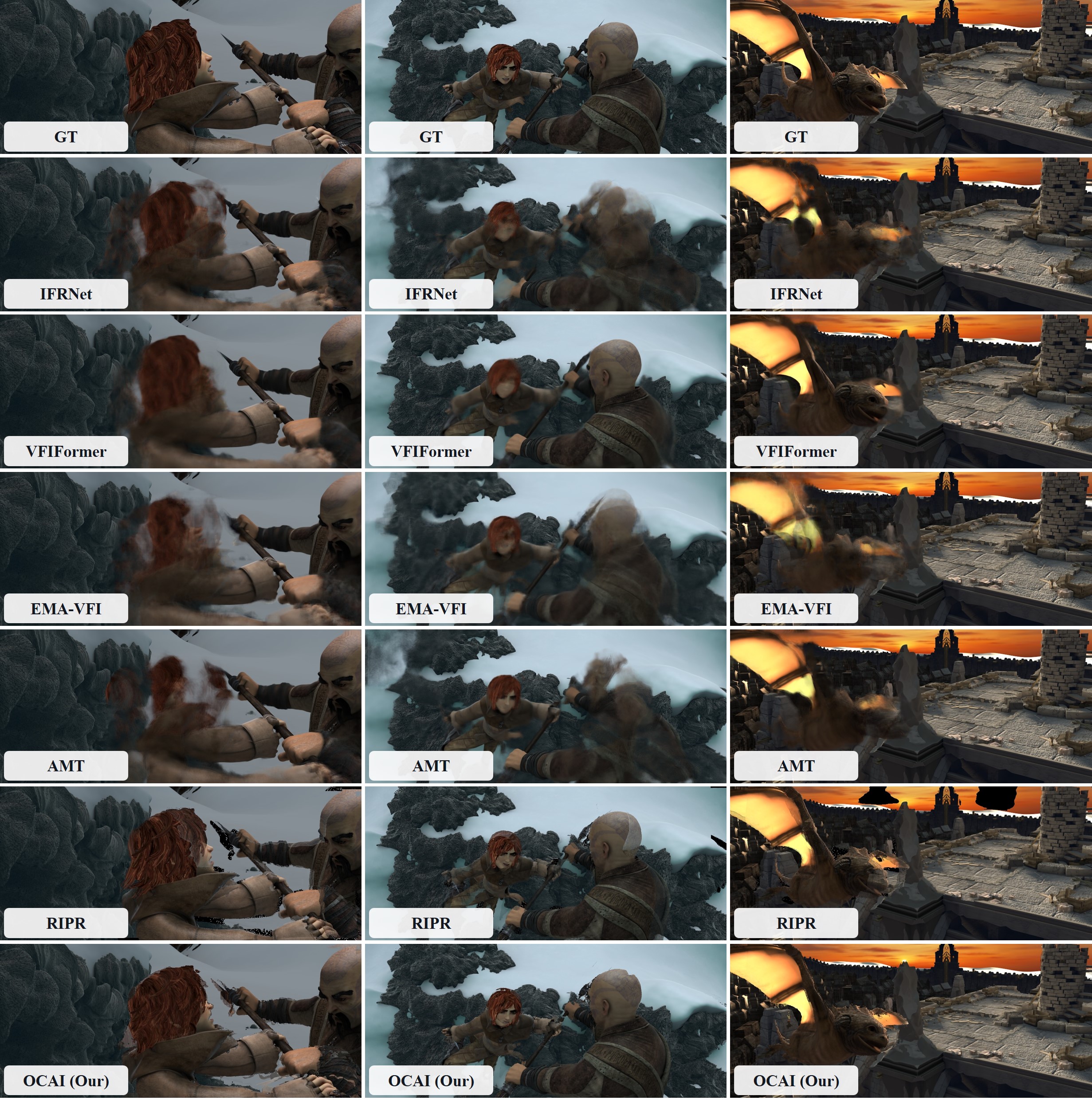}\\

\end{tabular}$
\end{center}
\vspace{-20pt}
\caption{\small Video Frame Interpolation (VFI) results on Sintel (clean). First row is the ground truth. Second to fifth rows are outputs of SOTA VFI models~\cite{kong2022ifrnet, lu2022video, zhang2023extracting, li2023amt}. Sixth row is the output of using RealFlow~\cite{han2022realflow} for VFI. Bottom row shows our OCAI results.  }
\label{exp:qualitative_vfi_sintel}
% \vspace{-10pt}
\vspace{-5pt}
\end{figure*}

\begin{figure*}[p]
% \begin{figure}[htbp]
\begin{center}$
\centering
\begin{tabular}{c}
\hspace{-0.3cm} \includegraphics[width=17.4cm]{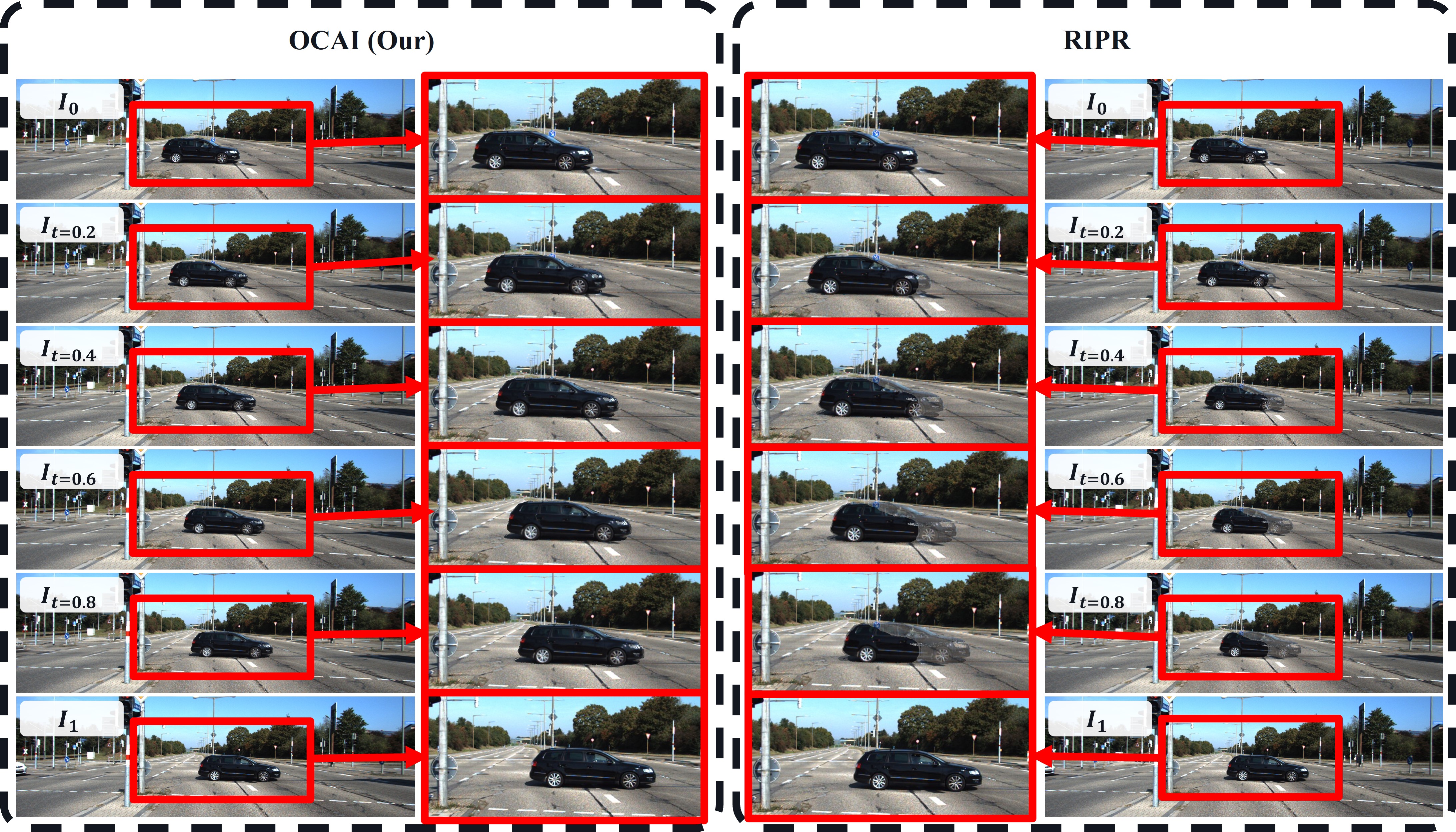}\\

\end{tabular}$
\end{center}
\vspace{-20pt}
\caption{\small Video Frame Interpolation (VFI) results on KITTI. We generate different $I_t$ images (for $t=0.2,\, 0.4,\, 0.6,\, 0.8$). Since backward warping cannot generate continuous inter-frames, we generate results using RIPR from RealFlow and our proposed OCAI approach.}
\label{exp:qualitative_vfi_kitti_time}
% \vspace{-10pt}
\vspace{-5pt}
\end{figure*}

\subsection{Optical Flow}
\textbf{Dataset. }
We follow semi-supervised optical flow training settings from previous work, e.g., FlowSupervisor~\cite{im2022semi}, RealFlow~\cite{han2022realflow}, and DistractFlow~\cite{jeong2023distractflow}.In Sintel test evaluation, we follow DistractFlow training pipeline and use Sintel training dataset and Monkaa dataset. In KITTI test evaluation, FlowSupervisor and DistractFlow use additional unlabeled datasets such as Driving and Spring, but RealFlow uses only KITTI multi-view training dataset. In our experiment, we follow RealFlow and use only KITTI multi-view training dataset. 

\textbf{Implementation Details. }
We follow FlowSupervisor, RealFlow, and DistractFlow settings. We set $\tau$ and $w$ as 0.95 and 1 in Eq.~13 and~14, same as in DistractFlow. We use initial decay rate in EMA of 0.99 and gradually increase it to 0.9996. Since our optical flow model already has been trained on C+T in a semi-supervised setting, we use a higher initial decay rate compared to \cite{liu2021unbiased} and use the same terminal decay rate as \cite{liu2021unbiased}.

\section{Additional Video Frame Interpolation results}

We generate more inter-frame images in Fig.~\ref{exp:qualitative_vfi_kitti},~\ref{exp:qualitative_vfi_sintel} on KITTI and Sintel datasets. In addition, we also generate more inter-frames with different t values (t = 0.2, 0.4, 0.6, 0.8). Since backward warping based VFI algorithms cannot generate continuous $I_{t}$ images, we compare inter-frames generated by our OCAI and RIPR from RealFlow in Fig.~\ref{exp:qualitative_vfi_kitti_time}. 

% \footnotetext{Note that PSNR and SSIM do not always correlate well with perceived visual quality, as shown in previous studies~\cite{zhang2018unreasonable, prashnani2018pieapp}. For instance, in Fig.~\ref{exp:qualitative_vfi}, VFIFormer has higher SSIM scores but worse visual quality.}

% \section{Rationale}
% \label{sec:rationale}
% % 
% Having the supplementary compiled together with the main paper means that:
% % 
% \begin{itemize}
% \item The supplementary can back-reference sections of the main paper, for example, we can refer to \cref{sec:intro};
% \item The main paper can forward reference sub-sections within the supplementary explicitly (e.g. referring to a particular experiment); 
% \item When submitted to arXiv, the supplementary will already included at the end of the paper.
% \end{itemize}
% % 
% To split the supplementary pages from the main paper, you can use \href{https://support.apple.com/en-ca/guide/preview/prvw11793/mac#:~:text=Delete%20a%20page%20from%20a,or%20choose%20Edit%20%3E%20Delete).}{Preview (on macOS)}, \href{https://www.adobe.com/acrobat/how-to/delete-pages-from-pdf.html#:~:text=Choose%20%E2%80%9CTools%E2%80%9D%20%3E%20%E2%80%9COrganize,or%20pages%20from%20the%20file.}{Adobe Acrobat} (on all OSs), as well as \href{https://superuser.com/questions/517986/is-it-possible-to-delete-some-pages-of-a-pdf-document}{command line tools}.


\end{document}